\documentclass{article}
\usepackage{iclr2027_conference,times}

\usepackage{amsmath,amsfonts,bm}

\def\eqref#1{equation~\ref{#1}}
\def\1{\bm{1}}

\DeclareMathAlphabet{\mathsfit}{\encodingdefault}{\sfdefault}{m}{sl}
\SetMathAlphabet{\mathsfit}{bold}{\encodingdefault}{\sfdefault}{bx}{n}

\usepackage{xcolor}
\usepackage{hyperref}
\hypersetup{
    colorlinks=true, 
    linkcolor=red,
    citecolor=gray, 
    urlcolor=magenta,
    pdfauthor={},
    pdfkeywords={}
}
\usepackage{url}
\usepackage{algorithm}
\usepackage{multirow}  % \multirow
\usepackage{graphicx}  % \resizebox
\usepackage{amsmath}
\usepackage{amssymb}
\usepackage{bm}
\usepackage{adjustbox}
\usepackage{booktabs}
\usepackage{array}
\usepackage{longtable}
\usepackage{tabularx}
\usepackage{enumitem}
\usepackage{algpseudocode}
\usepackage{listings}

\newcommand{\fullname}{VPTwin}
\newcommand{\modelname}{VPTwin-Core}

\title{%
\parbox{\textwidth}{\centering
VPTwin: Real-Sim-Real Video Prediction for\\
Robotic Manipulation Planning
}}

\author{
\makebox[\dimexpr\textwidth-2\tabcolsep\relax][c]{%
\begin{tabular}{@{}c@{}}
\noalign{\vskip 10pt}
Zhenghao Xiao$^{1}$, Minting Pan$^{2}$, Nantian He$^{3}$,
Dongzhan Zhou$^{2}$, Yunbo Wang$^{3\dagger}$ \\[6pt]
{\small\normalfont
$^{1}$Carnegie Mellon University \quad
$^{2}$Shanghai Artificial Intelligence Laboratory} \\[2pt]
{\small\normalfont
$^{3}$Shanghai Jiao Tong University} \\[4pt]
{\small\normalfont $^{\dagger}$Corresponding author}
\end{tabular}%
}
}

\iclrfinalcopy % Uncomment for camera-ready version, but NOT for submission.
\begin{document}

\maketitle
\lhead{} % Remove the conference-publication header
%\begin{abstract}
% than directly judging raw
% simulator rollouts.
%\end{abstract}

% Main sections: inserted in this order
\begin{abstract}

While action-conditioned video prediction provides an intuitive world model for robotics, purely data-driven predictors often suffer from compounding errors and physically implausible hallucinations in long-horizon rollouts, severely undermining downstream action planning. We propose VPTwin, a Real-Sim-Real video prediction framework that anchors real-world future prediction using real-synchronized simulation twins. For a target manipulation task, a VLM reconstructs an executable digital twin from a real demonstration episode. To accommodate the ill-posed estimation of unobserved physical properties, Isaac Sim simulates multiple forward dynamic rollouts across randomized physical configurations under candidate action trajectories. Using these rollouts as in-context references, VPTwin harmonizes both domains, using simulation dynamics to enforce physical plausibility while capturing unmodeled contact interactions from real video. Furthermore, we establish a predictive planning loop using VPTwin to visually verify VLM-proposed actions and guide reliable real-world execution. Evaluations show substantial reductions in physical hallucinations during video prediction and marked improvements in manipulation planning performance.

\end{abstract}
\section{Introduction}
\label{sec:Introduction}

Action-conditioned video prediction serves as an intuitive visual world model for robotic manipulation by allowing robots to imagine future visual states~\citep{wu2024ivideogpt, agarwal2026cosmos}.
However, purely data-driven predictors struggle with compounding errors during long-horizon autoregressive rollouts and often generate physically implausible hallucinations such as object interpenetration, floating artifacts, and spontaneous motion.
Crucially, hallucinations mislead downstream policy evaluation and cause action planning to fail.
Physics simulation offers a complementary remedy by enforcing explicit dynamical governing equations and conservation laws, guaranteeing physical consistency over extended horizons. Yet, relying solely on simulators inevitably exposes decision models to physical simulation-to-reality gaps in geometry, friction, and complex contacts~\citep{chen2024urdformer,liu2026simpact}.
Recognizing that simulation provides principled dynamic grounding while real-world video supplies unmodeled contact interactions and photorealistic fidelity, we propose a Real-Sim-Real video prediction paradigm that bridges both worlds.

\begin{figure}[!t]
\vspace{-5pt}
    \centering
    \includegraphics[width=\linewidth]{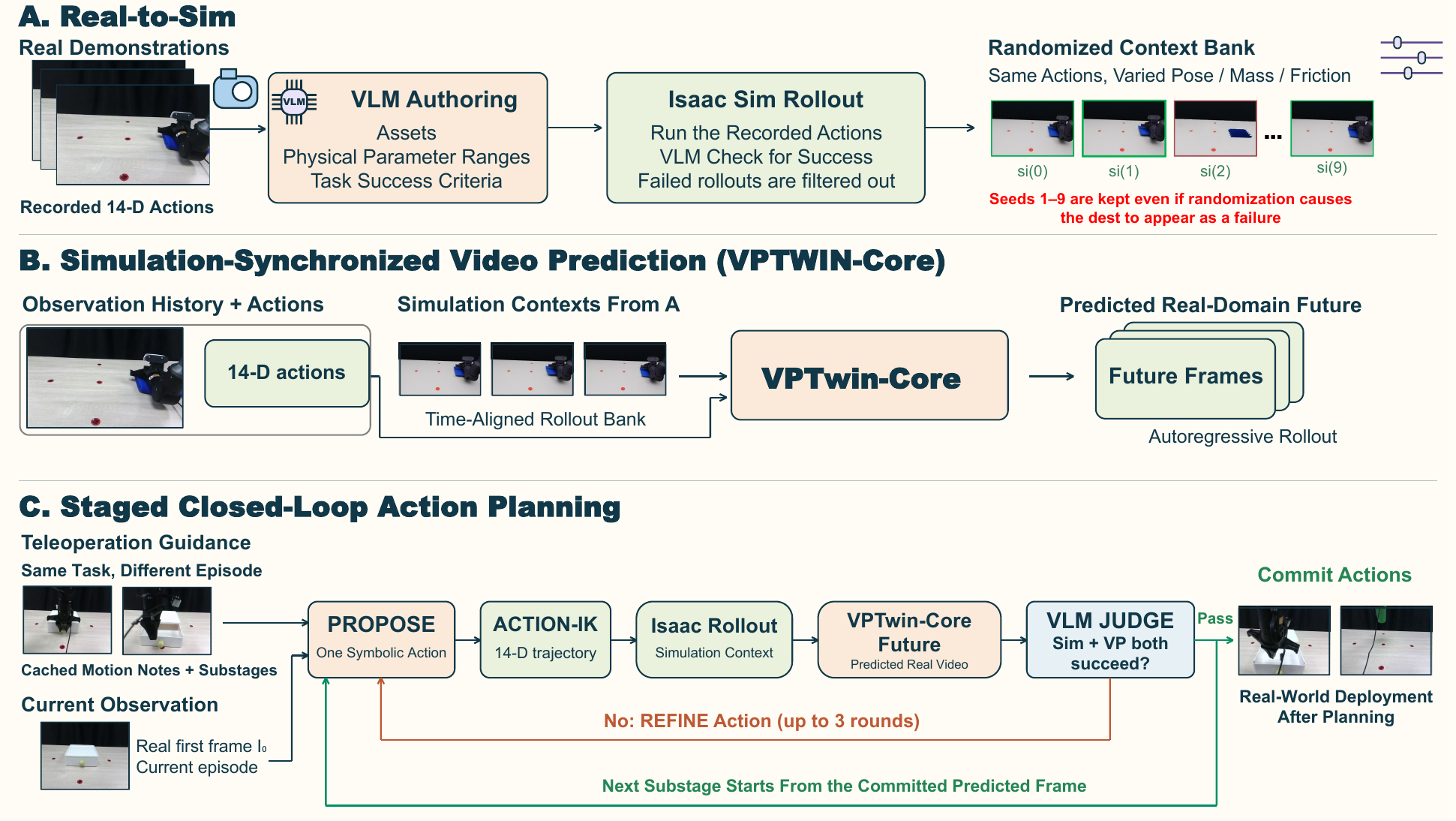}
    \vspace{-10pt}
    \caption{\textbf{Overview of \fullname{}.}
    (A) A VLM reconstructs executable digital-twin assets to drive randomized dynamic rollouts in Isaac Sim.
    (B) Conditioned on time-aligned simulation rollouts, \modelname{} synthesizes physically grounded real-domain future video under candidate actions.
    (C) Staged action planning evaluates and refines candidate actions via joint simulation and visual verification before real-world deployment.}
    \label{fig:vptwin-overview}
    % \vspace{-8pt}
\end{figure}

To realize this paradigm, we present VPTwin, which uses simulation dynamics to physically anchor real-domain video prediction for robotic manipulation.
Within this pipeline, a VLM reconstructs executable digital twins and estimates distributions over physical parameters directly from real-world scene observations.
Notably, frontier VLMs, such as GPT-6 Astra~\citep{openai2026gpt6astraapi}, substantially reduce the manual effort required to construct task-specific digital twins.
Since inferring exact, unobserved physical properties from visual observations is inherently ill-posed, we leverage Isaac Sim to simulate multiple forward dynamic rollouts under randomized physical parameter configurations, conditioned on the same future action trajectories (Figure~\ref{fig:vptwin-overview}A).

Building on these contexts, we introduce VPTwin-Core, a simulation-synchronized video prediction model that conditions on multiple possible future outcomes from the approximate digital twins, using them as physical references to mitigate physically implausible predictions and reduce forward hallucinations (Figure~\ref{fig:vptwin-overview}B).
Furthermore, to prevent over-optimistic predictions during planning, VPTwin incorporates failure-aware co-training: successful episodes are supervised by real-world demonstrations, while synthetic failure rollouts generated in simulation expose VPTwin-Core to counterfactual failure modes, without the need for costly real-world failure collection.

We establish a closed-loop planning framework that employs VPTwin-Core as a predictive visual verifier (Figure~\ref{fig:vptwin-overview}C).
Specifically, a VLM analyzes one-shot demonstrations drawn from the predictor's training set and decomposes them into semantically structured substages.
At each substage, the VLM proposes candidate actions, which are first evaluated in Isaac Sim and then projected into prospective real-world visual outcomes using VPTwin-Core.
An action is executed on the real robot only if its predicted visual outcome meets predefined success criteria, as assessed by the VLM. Otherwise, the predicted visual feedback guides iterative action refinement.
By screening candidate actions through physics-anchored visual foresight before real-world execution, the framework reduces hardware trial-and-error and helps prevent costly execution failures.

Our primary contributions are summarized as follows:
\begin{itemize}[leftmargin=*]
\vspace{-3pt}
\item We introduce VPTwin, a Real-Sim-Real vision-based model predictive control framework for robotic manipulation. By grounding generative visual rollouts in automated digital twins, it enforces physical plausibility throughout forward modeling.
\item We develop VPTwin-Core, an in-context world model conditioned on randomized, time-aligned physical rollouts, augmented by synthetic failure co-training to accurately foresee the visual futures under unsuccessful actions without collecting real failure data.
\item We design a staged planning algorithm that leverages VPTwin-Core as a predictive visual verifier to evaluate VLM-proposed actions and guide adaptive refinement prior to real-world execution.
\item Extensive evaluations across rigid, deformable, and articulated manipulation tasks demonstrate substantial reductions in physical hallucinations and marked gains in task success.
\vspace{-3pt}
\end{itemize}

% Maintaining physical consistency over long horizons remains a central challenge for action-conditioned video prediction. During autoregressive rollout, small prediction errors can accumulate into futures that remain visually plausible but no longer preserve object states or action outcomes~\citep{chen2026learning,wang2026interactive}. This matters for robot planning: a convincing depiction of task completion is useful only if the proposed action can produce it. Yet real-data-driven predictors offer an essential capability: they learn visual interaction dynamics directly from real experience, including contact and deformation that are difficult to specify accurately in a simulator~\citep{yang2023learning,wang2026interactive}. The challenge is therefore to improve long-horizon reliability while retaining this grounding in real-world interactions.

\section{Related Work}
\label{sec:related-work}

\vspace{-3pt}
\paragraph{Video prediction for robotic manipulation.}
Action-conditioned video prediction models the visual consequences of robot-object interactions, serving as an intuitive world model for manipulation planning and policy evaluation. Early frameworks such as UniPi convert generated video plans into robot actions~\citep{du2023learning}, while AVDC extracts control signals via dense visual correspondences and geometric reasoning~\citep{ko2024learning}.
Subsequent autoregressive and diffusion models learn controllable visual dynamics directly from demonstration data~\citep{wu2024ivideogpt,zhu2406irasim}. Recent efforts extend this paradigm to joint video-action modeling~\citep{li2025unifiedvideoactionmodel}, latent feature prediction for planning~\citep{zhou2024dino}, multiview generation~\citep{guo2026ctrl,team2025evaluating}, human video transfer~\citep{gao2026dreamdojo}, and motion-aligned visual control~\citep{cao2026tract,li2026hydra}
However, in contact-rich manipulation, visual history and actions cannot resolve unobserved physical parameters, causing extended predictions to accumulate physically implausible hallucinations~\citep{chen2026learning,wang2026interactive}. We therefore investigate whether alternative simulation rollouts across randomized physical parameters provide complementary context for predicting real-world continuations.

\vspace{-8pt}
\paragraph{Real-to-Sim scene reconstruction.}
Constructing interactive simulation environments from visual inputs is essential for physical modeling. 
Prior work reconstructs rigid, soft, and articulated assets from visual inputs~\citep{xiang2025structured,yang2024holodeck,zhang2026robosnap,huang2026soma,le2025articulate,chen2024urdformer,pfaff2026scenesmith}, with recent systems further automating executable URDF generation using LLM-based coding agents~\citep{zhou2026articraft,xiao2026beyond,he2026neoworld}.
While existing methods prioritize geometric and kinematic fidelity, identifying unobserved physical parameters from visual observation is inherently ill-posed. Rather than relying on a single deterministic estimate, our pipeline uses a VLM to infer plausible parameter distributions, synthesizing executable digital twins specifically designed to generate diverse physical rollout contexts.

\vspace{-8pt}
\paragraph{Simulation-guided video generation and planning.}
Modern physics engines provide controllable dynamic environments for embodied agents~\citep{todorov2012mujoco,nvidiaisaacsim,zhu2020robosuite}. Several works leverage simulation to facilitate visual learning: ReBot synthesizes robot training videos through simulation replay~\citep{fang2025rebot}, while RoboTransfer~\citep{liu2026robotransfer}, CRAFT~\citep{chen2026craft}, AnchorDream~\citep{ye2025anchordream}, and RealWonder~\citep{liu2026realwonder} guide video diffusion with simulated geometry or motion priors.
To mitigate dynamical discrepancy, PGRD learns residual corrections over simulated trajectories~\citep{patel2026learning}. 
\fullname{} departs from this by conditioning real-video prediction on multiple randomized rollouts of the identical action sequence, using simulation as a multi-hypothesis physical prior.
For downstream planning, frameworks such as Prompting-with-the-Future~\citep{ning2025prompting} and SIMPACT~\citep{liu2026simpact} optimize trajectories or symbolic plans within digital twins. Yet, planning entirely within simulation remains susceptible to physical reality gaps.
We address this with a staged Real-Sim-Real planner.

% Simulation platforms support controllable interaction~\citep{todorov2012mujoco,nvidiaisaacsim,zhu2020robosuite}.
% ReBot generates training videos through replay~\citep{fang2025rebot};
% RoboTransfer~\citep{liu2026robotransfer}, CRAFT~\citep{chen2026craft}, AnchorDream~\citep{ye2025anchordream}, and RealWonder~\citep{liu2026realwonder} guide video synthesis with geometry or simulated motion. PGRD learns corrections to simulated dynamics~\citep{patel2026learning}. Our question is whether several imperfect simulations of the \emph{same supplied action sequence} jointly improve prediction of its real outcome. We condition real-video prediction on multiple randomized replays of the same action sequence. For planning, Prompting-with-the-Future combines digital twins with VLM rewards~\citep{ning2025prompting}, while SIMPACT optimizes whole-skill symbolic plans in simulation~\citep{liu2026simpact}. Our staged planner uses a frozen fail-aware predictor alongside simulation, requiring both outcomes to satisfy each substage's success criterion (from VLM) before acceptance. On failure, REFINE revises that same action, at most three times, and the next substage starts from the committed predicted frame.

\section{Method}
\label{sec:method}

This section provides the technical details of \fullname{}. Sec.~\ref{sec:method-vlm} introduces our Real-to-Sim procedure for authoring executable digital twins from visual inputs under domain-randomized physics.
Sec.~\ref{sec:method-prediction} presents \modelname{}, a video prediction model conditioned on robot actions, visual history, and synchronized physics rollouts.
Sec.~\ref{sec:method-planning} establishes a planning algorithm that leverages the frozen predictive model to iteratively evaluate and refine action plans before physical execution.

\vspace{-3pt}
\subsection{Real-to-Sim Twin Authoring and Randomized Physics Rollouts}
\label{sec:method-vlm}
\vspace{-3pt}

To provide physics-grounded guidance for real-domain video prediction, we develop an automated Real-to-Sim pipeline that transforms raw demonstration videos into executable digital twins in Isaac Sim.
Because inferring exact, unobserved dynamic attributes (such as friction coefficients, mass distributions, and contact compliance) from passive visual observations is inherently ill-posed, treating a single simulated model as ground truth introduces severe model misspecification.
To overcome this fundamental limitation, our pipeline couples automated twin authoring with domain-randomized dynamic rollouts, generating a multi-hypothesis physical context that covers plausible dynamic variations rather than relying on brittle point estimates (see appendix Figure~\ref{fig:scene_authoring}).

\vspace{-8pt}
\paragraph{Automated digital-twin authoring.}
Given a real demonstration episode, we employ a frontier VLM (\textit{e.g.}, GPT-6 Astra)~\citep{openai2026gpt6astraapi} to parse third-view keyframes sampled across the trajectory ($0\%, 20\%, 40\%, 60\%, 70\%, 80\%, \text{and } 100\%$), identify object categories, detect geometric occlusions, infer kinematic articulation, and estimate initial placements on a calibrated workspace.
The VLM then programmatically generates Python scripts that instantiate Universal Scene Description (USD) assets, specifying collision primitives, inertial properties, and joint limits.
To guarantee physical stability within the physics engine, an automated coding agent~\citep{cursor2026grok46,spacexai2026grok46} refines actuator kinematics, converting the drawer as a fixed-base articulation with a prismatic joint and deploying deformable items as PhysX surface deformables.
In parallel, to benchmark our automated authoring against an expert baseline, we curate a measured-catalog setting consisting of $548$ real-world episodes across six task families, where object geometries are verified by human annotators.
Workspace calibration and asset construction are detailed in Appendix~\ref{app:experiments}, ~\ref{app:scene-assets}.

\vspace{-8pt}
\paragraph{VLM-driven success filtering.}
% Due to perceptual ambiguities in single-view reconstruction, reconstructed assets may occasionally experience kinematic divergence or unphysical instabilities during replay. 
To ensure that only physically valid simulations serve as predictive priors, the VLM establishes formal task specifications and visual success criteria for each task family (see appendix Table~\ref{tab:task_def}).
The nominal simulation rollout (seed~$0$), driven by recorded $14$-D actions, is evaluated via multi-timestamp visual inspection with denser temporal sampling during contact-rich phases. This automated verification retains $277$ physically consistent episodes across four manipulation families for downstream world-model training.

\vspace{-8pt}
\paragraph{Domain-randomized physical rollouts.}
For each retained demonstration, Isaac Sim rollouts the recorded $14$-D action sequence across a bank of ten randomized dynamic rollouts.
As shown in Figure \ref{fig:vptwin-overview}, seed~$0$ represents the nominal configuration, while seeds~$1$--$9$ systematically perturb object initial poses by $\Delta p$ and sample dynamical parameters $\vartheta$ (\textit{e.g.}, friction and mass) from VLM-specified plausible distributions, while holding the recorded action trajectory, tabletop geometry, camera calibration, and lighting invariant.
Rather than constraining the video predictor to an exact simulation twin, this ensemble of time-aligned rollouts exposes the model to diverse physical continuations under approximate digital twins.

% We construct two separate rollout sets. The measured-catalog setting covers 548 real episodes across six task families, using human-checked geometry and coding-agent-specified parameter ranges~\citep{cursor2026grok46,spacexai2026grok46}. The episodes were filtered by human.
% In the VLM-authored setting, GPT-6 Astra authors assets, success criteria, and parameter ranges (mass, friction, and others) from real-video stills~\citep{openai2026gpt6astraapi};A coding agent~\citep{cursor2026grok46, spacexai2026grok46} patches deployment so the drawer prismatic joint slides and both soft bodies work as PhysX surface deformables. A VLM success filter retains 277 episodes across four families. Both settings run recorded 14-D actions: seed~$0$ uses the accepted reference configuration, while seeds~$1$--$9$ vary
% placement and physics with actions and meshes fixed. The settings do not share simulation conditions. Figure~\ref{fig:vptwin-overview}(A) summarizes the pipeline; Appendix~\ref{app:scene-assets} details asset construction. Since real demonstrations contain only successes, we perturb actions to construct failures for \textit{Grasp Cup} and \textit{Open and Close Drawer} (Appendix~\ref{app:fail-dests}).

\vspace{-3pt}
\subsection{Simulation-Synchronized Video Prediction}
\label{sec:method-prediction}
\vspace{-3pt}

\paragraph{Counterfactual failure synthesis.}
Standard teleoperation datasets are often positively biased, containing almost exclusively successful executions. As a result, video predictors may develop overly optimistic predictions, hallucinating successful outcomes even for flawed candidate actions.
To reliably verify downstream plans, the predictive model must generalize to counterfactual actions and faithfully visualize failure outcomes. Because collecting physical failures on hardware is costly and hazardous, we synthesize failure trajectories in simulation.
For episode $i$, let $r_{i,t}$ and $a_{i,t}\in\mathbb{R}^{14}$ denote the visual observation and joint action, defining the full trajectory $A_i=(a_{i,0},\ldots,a_{i,T_i-1})$. While successful demonstrations use real video supervision, failure episodes apply kinematic perturbations during critical phases (\textit{e.g.}, misaligned approaches or premature gripper releases; Appendix~\ref{app:fail-dests}). 
The nominal simulation rollout $s_i^{(0)}$ under perturbed actions $A_i$ then serves as the pseudo-observation target, providing failure supervision without collecting real failure data.
%
% For failure-aware co-training, we curate a balanced dataset pairing 118 successful real demonstration episodes drawn from the filtered 277-episode VLM set with 118 synthesized failure counterparts.

\begin{figure}[!t]
\vspace{-5pt}
    \centering
    \includegraphics[width=\linewidth]{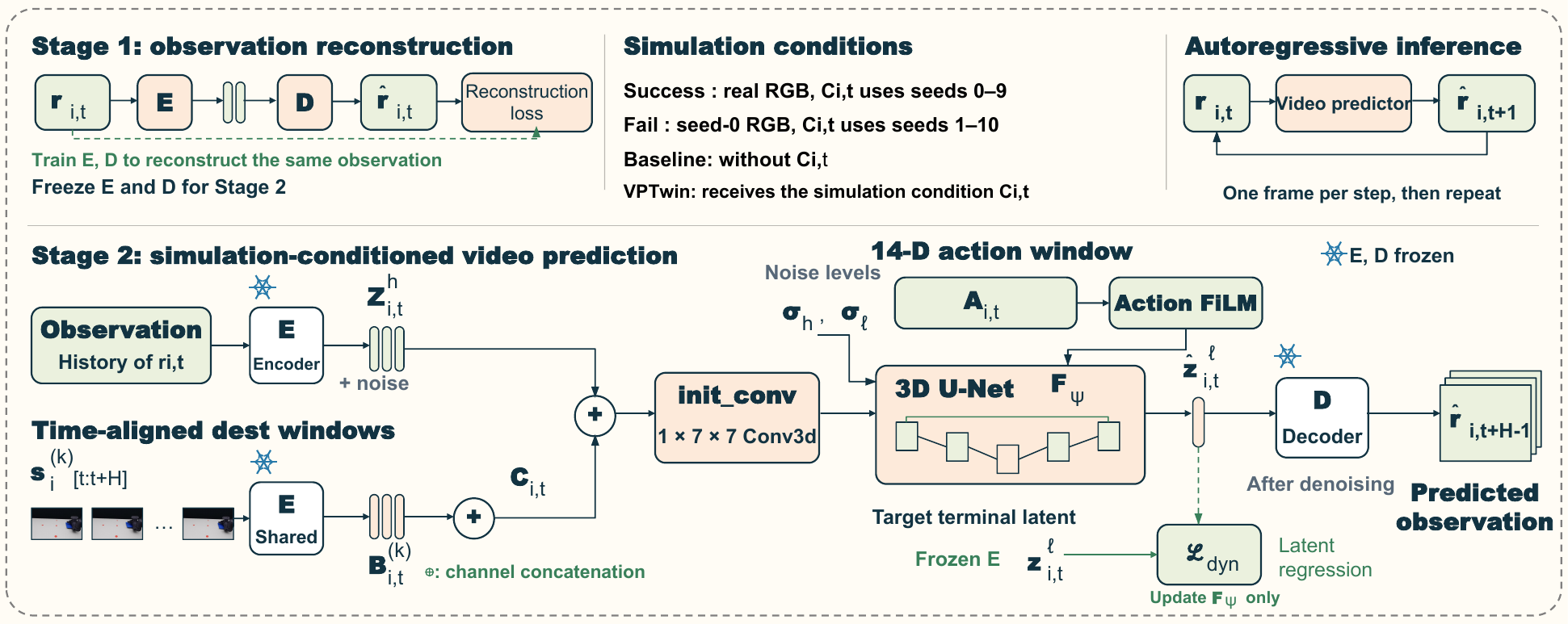}
    \vspace{-10pt}
    \caption{\textbf{Architecture of \modelname{}.} Two-stage training and autoregressive inference conditioned on candidate actions and time-aligned simulation rollouts.}
    \label{fig:predictor-details}
    % \vspace{-8pt}
\end{figure}

\vspace{-8pt}
\paragraph{\modelname{} architecture.}

Given the co-training dataset, Isaac Sim simulates forward dynamic rollouts driven by action sequence $A_i$ across varied physical parameters and initial placements to generate a multi-hypothesis physical context pool:
% For episode $i$, let $r_{i,t}$ denote the observation RGB at time $t$ and $a_{i,t}\in\mathbb{R}^{14}$ the corresponding joint action. We write the complete action trajectory as $A_i=(a_{i,0},\ldots,a_{i,T_i-1})$ and use $A_{i,t}=A_i[t:t+H]$ for the action window aligned with a Stage-2 sample. Isaac produces a pool of dest videos
\begin{equation}
s_i^{(k)}
=
\mathcal{R}\!\left(
A_i,
p_i+\Delta p_{i,k},
\vartheta_{i,k};
\Gamma
\right),
\qquad
k\in\mathcal{K}_i,
\label{eq:rollout-pool}
\end{equation}
where $p_i$ is the nominal object placement, $\Delta p{i,k}$ represents spatial placement perturbation, $\vartheta_{i,k}$ denotes sampled physical parameters, and $\Gamma$ specifies static scene parameters, including tabletop geometry, camera poses, lighting, and robot base calibration.
The rollout configuration index set is $\mathcal{K}_i=\{0,\ldots,9\}$ for successful episodes and $\mathcal{K}_i=\{0,\ldots,10\}$ for counterfactual failure episodes.
Given a prediction horizon $H=10$, let $A_{i,t}=A_i[t:t+H]$ define the time-aligned action window. The visual latent representation of the $k$-th simulation rollout window is extracted via a frozen encoder as $B_{i,t}^{(k)} = E(s_i^{(k)}[t:t+H])$.
As shown in Figure~\ref{fig:predictor-details}, following the two-stage formulation of IWS~\citep{wang2026interactive}, Stage~1 trains a CNN encoder $E$ and a consistency decoder $D$ to reconstruct RGB frames via latent-conditioned denoising. 
Stage~2 freezes both $E$ and $D$, and trains a 3D spatiotemporal dynamics model $F_\psi$.
Observation, action, and simulation rollout windows are strictly time-aligned.
% where $p_i$ is the reference object placement, $\Delta p_{i,k}$ is the placement perturbation, $\vartheta_{i,k}$ contains the sampled physical parameters, and $\Gamma$ contains the fixed table, cameras, lights, and robot. The pool indices are $\mathcal{K}_i=\{0,\ldots,9\}$ for success episodes and
% $\mathcal{K}_i=\{0,\ldots,10\}$ for fail episodes. For the Stage-2 horizon $H=10$, the latent clip of the $k$th time-aligned dest window is
% \begin{equation}
% B_{i,t}^{(k)}
% =
% E\!\left(s_i^{(k)}[t:t+H]\right).
% \label{eq:dest-latent}
% \end{equation}
% Following IWS~\citep{wang2026interactive}, Stage~1 trains a CNN~\citep{lecun1998gradient} encoder $E$ and a consistency decoder $D$ to reconstruct RGB through latent-conditioned denoising. Stage~2 freezes both and trains $F_\psi$ (Figure~\ref{fig:predictor-details}). Observation, action, and rollout windows are time-aligned (Appendix~\ref{app:training}). 
Let $Z_{i,t}^{h}$ denote the mixed-noise latent clip of $r_i[t:t+H]$, with $H-1$ lightly noised history slots and terminal noise level $\sigma_h$. The target $z_{i,t}^{\ell}$ is the latent of $r_{i,t+H-1}$ at noise level $\sigma_\ell$. We concatenate ten rollout latent clips:
\begin{equation}
C_{i,t}
=
\oplus_{k\in\mathcal{I}_i} B_{i,t}^{(k)},
\qquad
\mathcal{I}_i
=
\begin{cases}
\{0,\ldots,9\}, & \text{success episode},\\
\{1,\ldots,10\}, & \text{fail episode},
\end{cases}
\label{eq:simulation-synchronized}
\end{equation}
where $\oplus$ denotes channel concatenation. The dynamics model
predicts the terminal low-noise latent:
\begin{equation}
\widehat z_{i,t}^{\ell}
=
F_\psi\!\left(
Z_{i,t}^{h}\oplus C_{i,t};
A_{i,t},
\sigma_h,
\sigma_\ell
\right).
\label{eq:conditioned-dynamics}
\end{equation}
In Stage~2, the same frozen encoder $E$ encodes observations and simulation rollouts. The 3D U-Net $F_\psi$ receives the concatenated volume $Z_{i,t}^{h}\oplus C_{i,t}$ at its first convolution and modulates intermediate spatiotemporal features with the action window $A_{i,t}$ via Feature-wise Linear Modulation (FiLM)~\citep{perez2018film}.
During autoregressive inference, each predicted latent step updates the sliding context window to condition subsequent steps before being decoded into RGB video frames.
%
% In Figure~\ref{fig:predictor-details}, $\tau$ denotes the current inference time, whereas $t$ indexes the training-window start.
The dynamics network is optimized under a uniform-MSE objective: $\mathcal{L}_{\mathrm{dyn}}(\psi)
=
\mathbb{E}[
\|
\widehat z_{i,t}^{\ell}
-
z_{i,t}^{\ell}
\|_2^2
].$
% This objective supervises the terminal observation latent. 
The exact noise schedule and effective terminal-only regression are detailed in Appendix~\ref{app:training}. 
% For successful episodes in both settings, the observation window and prediction target come from paired real video; simulated rollouts enter only through $C_{i,t}$.  For failure examples, $A_i$ denotes the perturbed action trajectory used to generate both the observation rollouts and conditioning rollouts. The observation window is
% \begin{equation}
% r_i[t:t+H]
% =
% s_i^{(0)}[t:t+H],
% \label{eq:fail-observation}
% \end{equation}
% so the seed-$0$ terminal frame supplies the target, while seeds~$1$--$10$ form $C_{i,t}$. For fail-aware co-training, we pair 118 successful real episodes drawn from the filtered 277-episode VLM set with their 118 constructed fail counterparts. Stage~1 reconstructs real success frames together with seed-$0$ fail frames; seeds~$1$--$10$ are not reconstruction targets. Stage~2 jointly trains the success and fail cases defined in
% Eq.~\eqref{eq:simulation-synchronized}. The no-simulation baseline removes $C_{i,t}$ while retaining
% the same observation input, frozen encoder and decoder, data split, action windows, and training budget.

\vspace{-3pt}
\subsection{\fullname{}-Based Model Predictive Control}
\label{sec:method-planning}
\vspace{-3pt}
To translate physics-grounded predictive foresight into reliable real-world manipulation, we formulate a staged Real-Sim-Real closed-loop planning framework that employs the frozen \modelname{} model as a predictive visual verifier.
Directly optimizing symbolic plans across an entire long-horizon skill frequently suffers from compounding search errors. To mitigate this, we decompose complex manipulation skills into semantically structured substages informed by demonstration motion priors.
Specifically, a VLM first analyzes teleoperation keyframes from the training set to construct a structured motion specification per task family, defining sequential 3D Cartesian waypoints, end-effector wrist orientations, gripper actuation timings, and critical contact milestones.
A secondary pass establishes substage boundaries, permissible motion primitives, and local success criteria. 
The planner comprises five steps: scene reconstruction is performed once, followed by steps 2--5 within each substage. The complete trajectory is executed after all substages are planned (Algorithm~\ref{alg:staged-planning} in Appendix~\ref{app:action_planning} details the procedure).
\begin{enumerate}[leftmargin=*]
\vspace{-3pt}
\item \textbf{Reconstruction:} Identifies the target task family and spatial workspace configuration directly from the initial real observation frame $I_0$, once before substage planning.
\item \textbf{Propose:} Prompts a VLM to generate a candidate symbolic action conditioned on the current visual observation, reference demonstration keyframes, and the nominal simulated state.
\item \textbf{Action-IK:} Converts symbolic parameters (translations, wrist rotations, and gripper commands) into an executable 14-D joint trajectory via inverse kinematics~\citep{lynch2017modern} with $2\,\mathrm{cm}$ Cartesian interpolation. Isaac Sim subsequently simulates forward dynamic rollouts across seeds~$0$--$9$ to construct the physical conditioning context for visual prediction.
\item \textbf{Judge:} Evaluates the proposed action against the substage success criterion using both the nominal simulation rollout (seed~$0$) and the real-domain visual future synthesized by \modelname{}. Crucially, simulation-only checks leave the planner vulnerable to unmodeled discrepancies in contact and friction, leading to false acceptances during hardware deployment. In contrast, our dual-domain verification helps prevent physically infeasible actions from being deployed.
\item \textbf{Refine \& Commit:} If either the simulation rollout or the predicted real future fails the check, an adaptive refinement module prompts the VLM to revise the action using the visualized failure rollout, demonstrations, and historical feedback for up to three rounds; detailed in Appendix~\ref{app:action_planning}. The first passing candidate is committed, and its final
predicted frame and simulator state initialize the next substage. If none passes, planning terminates without hardware execution.
%Upon acceptance, the action is deployed on the real robot, and the terminal predicted visual frame and simulator state initialize the subsequent substage.
\vspace{-3pt}
\end{enumerate}

    \section{Experiment}
    \label{sec:experiment}
    
    \begin{table}[t]
    \vspace{-5pt}
        \centering
        \caption{\textbf{Full-episode video prediction results} (mean $\pm$ std across tasks after averaging each family's scores over five training seeds)
        %(mean $\pm$ std over $5$ training seeds).
        % Baseline is action-only; Ours uses 10 simulation rollouts.
        % Bold indicates the better mean.
        Corresponding results for each task are presented in Table~\ref{tab:full_episode_per_family}.}
        \label{tab:full_episode_avg}
        \setlength{\tabcolsep}{3pt}
        \resizebox{\linewidth}{!}{%
        \begin{tabular}{@{}lccccccc@{}}
            \toprule
            \textbf{Model}
            & MSE $\downarrow$
            & LPIPS $\downarrow$
            & FID $\downarrow$
            & FVD $\downarrow$
            & PSNR $\uparrow$
            & SSIM $\uparrow$
            & UIQI $\uparrow$ \\
            \midrule
            \multicolumn{8}{l}{\textit{Measured-catalog (548), 6 families}} \\
            IWS \citeyearpar{wang2026interactive}
            & $0.016 \pm 0.008$
            & $0.329 \pm 0.071$
            & $276.0 \pm 32.1$
            & $2095.0 \pm 384.4$
            & $25.58 \pm 3.00$
            & $0.770 \pm 0.069$
            & $0.249 \pm 0.049$ \\
            \fullname{}
            & $\mathbf{0.007 \pm 0.002}$
            & $\mathbf{0.284 \pm 0.056}$
            & $\mathbf{265.6 \pm 36.9}$
            & $\mathbf{2014.5 \pm 264.7}$
            & $\mathbf{28.57 \pm 1.92}$
            & $\mathbf{0.798 \pm 0.058}$
            & $\mathbf{0.283 \pm 0.040}$ \\
            \midrule
            \multicolumn{8}{l}{\textit{VLM-authored (277), 4 families}} \\
            IWS \citeyearpar{wang2026interactive}
            & $0.016 \pm 0.009$
            & $0.310 \pm 0.067$
            & $277.7 \pm 45.3$
            & $2450.0 \pm 701.8$
            & $25.64 \pm 3.35$
            & $0.780 \pm 0.073$
            & $0.262 \pm 0.048$ \\
            \fullname{}
            & $\mathbf{0.006 \pm 0.002}$
            & $\mathbf{0.259 \pm 0.048}$
            & $\mathbf{259.1 \pm 42.3}$
            & $\mathbf{2131.0 \pm 340.6}$
            & $\mathbf{28.96 \pm 1.83}$
            & $\mathbf{0.809 \pm 0.060}$
            & $\mathbf{0.294 \pm 0.039}$ \\
            \bottomrule
        \end{tabular}%
        }
        % \vspace{-8pt}
    \end{table}
    
    \begin{figure}[t]
        \centering
        \includegraphics[width=\linewidth]{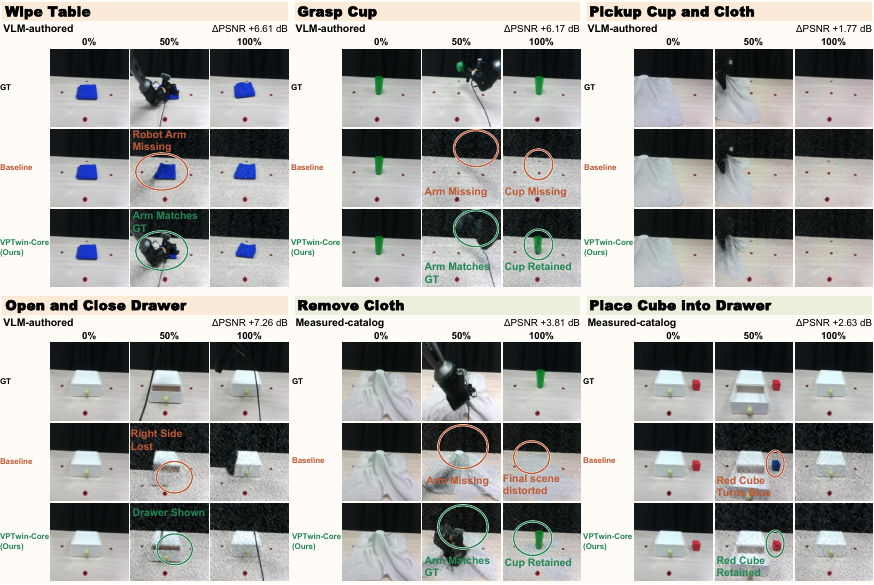}
        \vspace{-10pt}
        \caption{\textbf{Full-episode video prediction results at $0\%$, $50\%$, and $100\%$ of an execution timeline.}
        Orange and green circles highlight baseline failure artifacts and our physically consistent predictions, respectively. $\Delta\mathrm{PSNR}$ denotes the quantitative gain of \fullname{} over IWS~\citep{wang2026interactive}.}
        \label{fig:5}
        % \vspace{-5pt}
    \end{figure}
    
    \vspace{-3pt}
    \subsection{Experimental Setup}
    \vspace{-3pt}
    % \paragraph{Benchmarks.}
    We evaluate visual prediction on held-out episodes across two distinct digital-twin settings: (i) the \textit{measured-catalog} setting ($548$ episodes across six tasks) and (ii) the automated \textit{VLM-authored} setting ($277$ episodes across four tasks).
    The suite spans six tabletop tasks:
    \textit{Wipe Table} (surface wiping with a cloth remaining on the table),
    \textit{Grasp Cup} (lifting and setting a cup upright near its start, or at another slot in measured-catalog variants),
    \textit{Pickup Cup and Cloth} (bimanual lifting of both objects out of view),
    \textit{Remove Cloth} (clearing the cloth while leaving the cup upright),
    \textit{Open and Close Drawer} (sequential prismatic drawer articulation), and
    \textit{Place Cube into Drawer} (opening a drawer, inserting a cube, and closing it).
    % The VLM-authored benchmark evaluates the first three tasks alongside \textit{Open and Close Drawer}.
    %
    % \vspace{-8pt}
    % \paragraph{Robot deployment setup.}
    We benchmark real-world planning against SIMPACT~\citep{liu2026simpact} and open-loop $\pi_{0.5}$~\citep{intelligence2025pi_} on \textit{Grasp Cup} ($7$ trials, rigid-body contact, five placement slots) and \textit{Open and Close Drawer} ($5$ trials, articulated constraint, centered). 
    To assess execution robustness under physical uncertainty, object placements are perturbed across episodes.
    Experiments use bimanual ALOHA with AgileX Piper arms and a static third-person camera. (Appendix Figure~\ref{fig:experiment-setup}).
    Each trial initiates from frame $I_0$ and executes a committed $14$-D joint trajectory. Teleoperation demonstrations are a separate training set. Details are in Appendix~\ref{app:experiments}.
    
    Following IWS~\citep{wang2026interactive}, we evaluate MSE/PSNR~\citep{wang2009mean}, SSIM~\citep{wang2004image}, UIQI~\citep{wang2002universal}, LPIPS~\citep{zhang2018unreasonable}, FID~\citep{heusel2017gans}, and FVD~\citep{unterthiner2018towards} using checkpoints trained for $10^6$ steps.

\vspace{-3pt}
\subsection{Long-Term Video Prediction Results}
\vspace{-3pt}
As shown in Table~\ref{tab:full_episode_avg}, \fullname{} consistently outperforms IWS across all family-averaged metrics.
It improves $\mathrm{PSNR}$ by $2.99\,\mathrm{dB}$ in the measured-catalog setting and by $3.32\,\mathrm{dB}$ in the VLM-authored setting, while reducing $\mathrm{LPIPS}$ from $0.329$ to $0.284$ and from $0.310$ to $0.259$.
Figure~\ref{fig:5} shows that our model substantially mitigates temporal drift, accurately tracking manipulator articulation, object permanence (\textit{e.g.}, cup presence), and rigid drawer geometry throughout extended rollouts.

% VPTWin-Core improves all family-averaged metrics (Table~\ref{tab:full_episode_avg}), increasing PSNR by
% $2.99$/$3.32$\,dB and reduces LPIPS from $0.329$/$0.310$ to $0.284$/$0.259$ in the measured-catalog/VLM-authored settings, respectively. Figure~\ref{fig:5} shows better preservation of arm position,
% cup presence, drawer geometry, and cube color. Differences are subtler for VLM-authored
% \textit{Pickup Cup and Cloth}. These comparisons assess visual fidelity, not physical correctness.

\begin{table}[t]
\centering
\caption{\textbf{Real-robot task success and critic evaluation metrics.} Precision, Recall, and F1 assess critic verification against true physical outcomes: Precision denotes physical success rates among accepted plans, while Recall reflects the proportion of successful executions accepted. Rejected plans are additionally executed for evaluation.}
\label{tab:real_critic}
\small
\begin{tabular}{lcccc cccc}
\toprule
& \multicolumn{4}{c}{\textbf{Grasp Cup}} & \multicolumn{4}{c}{\textbf{Open and Close Drawer}} \\
\cmidrule(lr){2-5} \cmidrule(lr){6-9}
\textbf{Method} & Success & Precision & Recall & F1-score & Success & Precision & Recall & F1-score \\
\midrule
$\pi_{0.5}$ & 0/7 & na & na & na & 0/5 & na & na & na \\
SIMPACT & 2/7 & 0.33 & 1.00 & 0.50 & 2/5 & 0.67 & 1.00 & 0.80 \\
VPTwin & \textbf{3/7} & \textbf{1.00} & \textbf{1.00} & \textbf{1.00} & \textbf{5/5} & \textbf{1.00} & \textbf{1.00} & \textbf{1.00} \\
\bottomrule
\end{tabular}
\end{table}

\begin{figure}[!t]
    \centering
    \includegraphics[width=\linewidth]{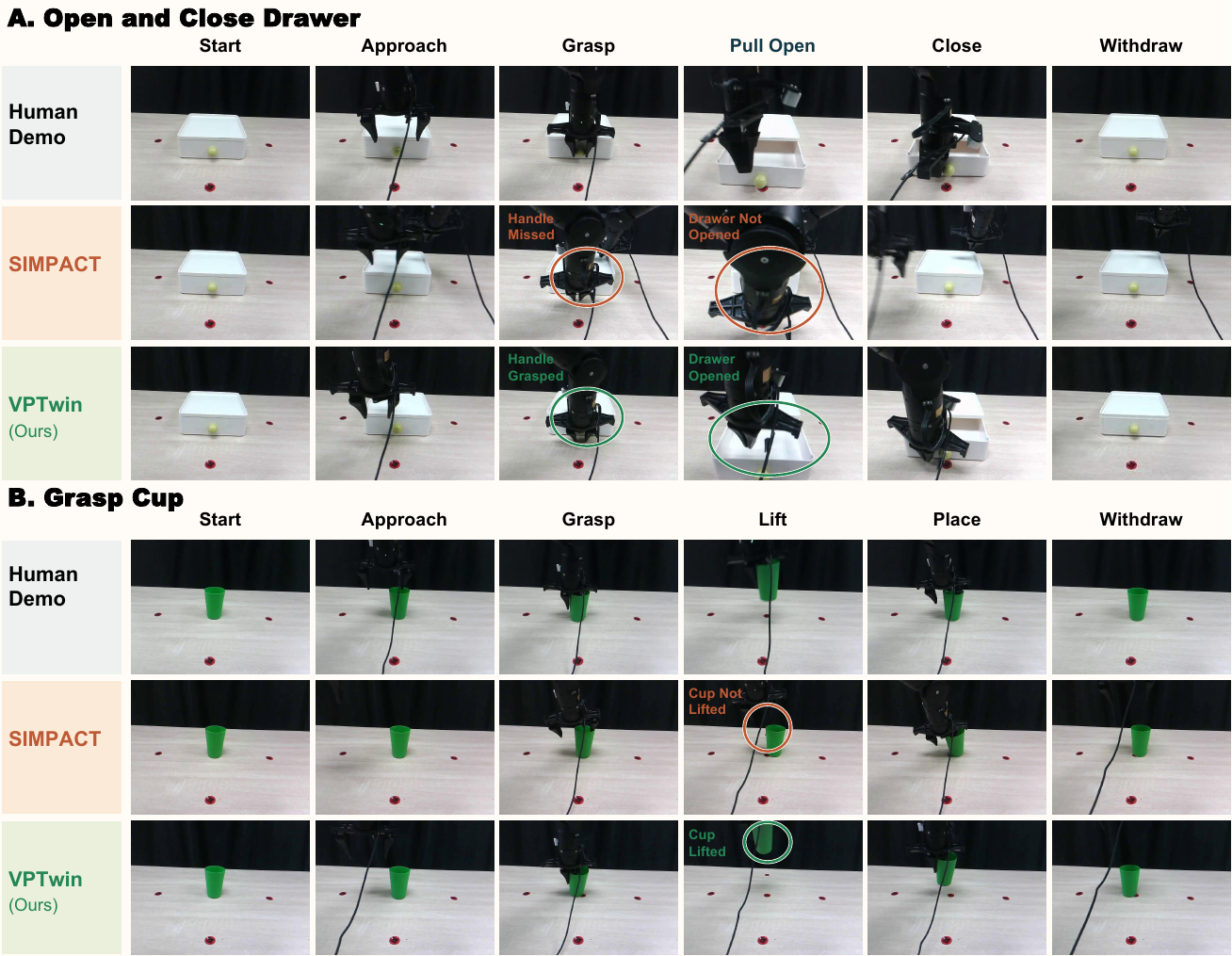}
    \vspace{-10pt}
    \caption{\textbf{Real-robot deployment trajectories.}
    % Human demonstrations and hardware executions of SIMPACT
    % and VPTwin for (A) \textit{Open and Close Drawer} and
    % (B) \textit{Grasp Cup}.
    Columns represent sequential task phases. Orange and green circles highlight failure and success modes, respectively.
    }
    \label{fig:7}
    % \vspace{-5pt}
\end{figure}

\vspace{-3pt}
\subsection{Action Planning Results}
\vspace{-3pt}

As reported in Table~\ref{tab:real_critic}, VPTwin demonstrates superior real-world manipulation performance over the baselines, improving execution success over SIMPACT from $2/7$ to $3/7$ on \textit{Grasp Cup} and achieving a perfect $5/5$ (versus $2/5$) on \textit{Open and Close Drawer}, while open-loop $\pi_{0.5}$ fails across all trials.
Crucially, evaluating the alignment between critic decisions and physical execution outcomes reveals the key vulnerability of simulation-only planning: while both planners achieve a Recall of $1.00$, SIMPACT's simulation critic incurs five false acceptances (four in grasping and one in drawer articulation), yielding low Precision scores of $0.33$ and $0.67$. In contrast, our dual-domain visual verification strictly rejects physically infeasible candidates, achieving perfect Precision and $\mathrm{F}_1$ scores of $1.00$ across both tasks by completely eliminating false acceptances.
Figure~\ref{fig:7} illustrates SIMPACT missing the drawer handle and failing to lift the cup, while VPTwin completes both tasks.

\vspace{-3pt}
\subsection{Ablation Studies}
\vspace{-3pt}
\paragraph{Effect of synthetic failure supervision.}
Incorporating simulated failure rollouts consistently improves visual prediction over extended execution horizons (Table~\ref{tab:failmix_277}). For \textit{Open and Close Drawer}, full-rollout $\mathrm{PSNR}$ increases by $0.97\,\mathrm{dB}$ (first-frame initialization) and $0.86\,\mathrm{dB}$ (perturbed-frame initialization), compared to $0.07\,\mathrm{dB}$ and $0.29\,\mathrm{dB}$ across $30$-frame rollouts. Full-rollout gains for \textit{Grasp Cup} reach $0.08$--$0.09\,\mathrm{dB}$, with consistent gains reflected in $\mathrm{SSIM}$ (Table~\ref{tab:failmix_277_ssim}). Co-training successfully prevents optimistic hallucinations, faithfully visualizing the incomplete drawer state and anticipating cup slippage under flawed actions where the success-only baseline fails.

\begin{figure}[t]
    \centering
    \includegraphics[width=\linewidth]{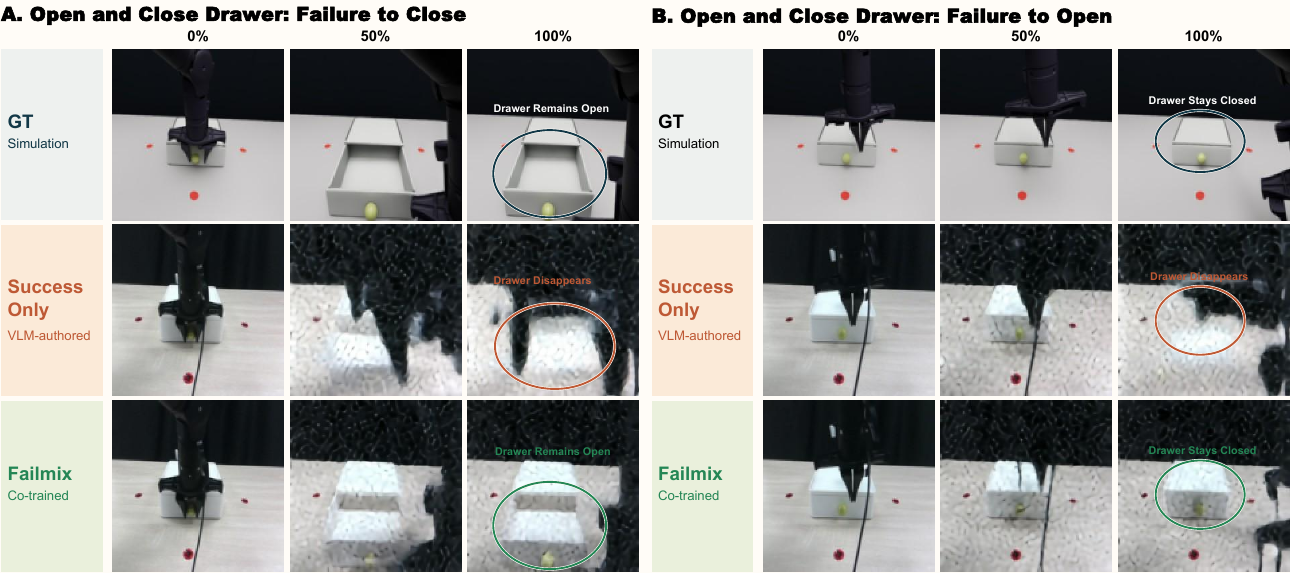}
    \vspace{-10pt}
    \caption{\textbf{Visual prediction rollouts on failure episodes.}
    Frame sequences across simulation ground truth, the success-only model, and the failure-aware co-trained model (\textit{Failmix}) at $0\%$, $50\%$, and $100\%$ completion for (A) unachieved drawer closing and (B) unachieved drawer opening. }
    \label{fig:6}
    \vspace{-5pt}
\end{figure}

\begin{table}[!t]
\centering
\caption{\textbf{Prediction results of failure episodes in PSNR (dB).}
% Performance across $3$ \textit{Grasp Cup} and $8$ \textit{Open and Close Drawer} test trajectories. 
\textit{Failmix} incorporates synthetic failure co-training, while \textit{Success-Only} relies solely on the $277$ filtered success demonstrations. Scores are evaluated on $30$-frame prefixes and full rollouts.}
\label{tab:failmix_277}
\small
\setlength{\tabcolsep}{8pt}
\begin{tabular}{@{}llcccc@{}}
\toprule
& & \multicolumn{2}{c}{30 frames} & \multicolumn{2}{c}{Full-episode} \\
\cmidrule(lr){3-4} \cmidrule(lr){5-6}
Start & Task & \textbf{Failmix} & \textbf{Success-only}
& \textbf{Failmix} & \textbf{Success-only} \\
\midrule
\multirow{2}{*}{First frame}
& Grasp Cup & \textbf{23.38} & 23.30 & \textbf{23.21} & 23.13 \\
& Open and Close Drawer & \textbf{22.46} & 22.39 & \textbf{21.22} & 20.25 \\
\midrule
\multirow{2}{*}{Perturbed frame}
& Grasp Cup & \textbf{23.36} & 23.23 & \textbf{23.20} & 23.11 \\
& Open and Close Drawer & \textbf{21.32} & 21.03 & \textbf{20.76} & 19.90 \\
\bottomrule
\end{tabular}
\end{table}

\begin{figure*}[t]
\vspace{-8pt}
    \centering
    \includegraphics[width=\textwidth]{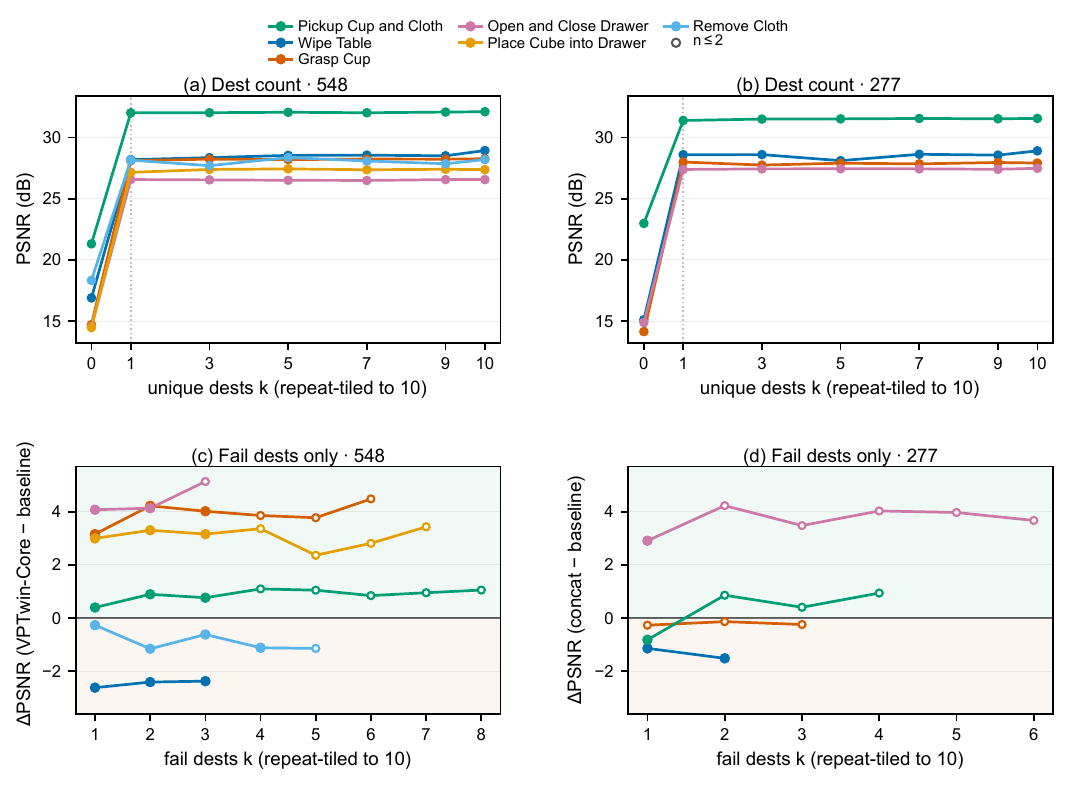}
    \vspace{-20pt}
    \caption{\textbf{Sensitivity to context quantity and outcome mismatch during inference.}
    Columns distinguish digital-twin settings: (a, c) measured-catalog and (b, d) VLM-authored.
    \textbf{(a, b)} Full-test $\mathrm{PSNR}$ as a function of the number of distinct simulation rollouts $k$ cyclically repeated across the 10 conditioning slots ($k=0$ corresponds to zero simulation context); values for $k\in\{1,3,5,7,9\}$ report averaged results across three evaluation seeds.
    \textbf{(c, d)} Relative prediction gain under counterfactual, failure-only simulation context. Open markers denote sparse sample regimes ($n \le 2$).}
    \label{fig:dest_ablation}
    % \vspace{-5pt}
\end{figure*}

\vspace{-8pt}
\paragraph{Sensitivity to context volume.}
Figure~\ref{fig:dest_ablation}(a,b) evaluates model sensitivity to the number of simulation rollouts provided at inference.
For models trained with ten rollouts, one test-time rollout ($k=1$) substantially improves PSNR over zero context ($k=0$).
However, this inference-time plateau does not imply that fewer rollouts suffice during training. 
As shown in Table~\ref{tab:context3_vs10_psnr}, when evaluated with an identical budget of three test-time rollouts, the model trained with ten rollouts outperforms the three-rollout training configuration across all six measured-catalog tasks.
Full results are presented in Appendix Tables~\ref{tab:context3_vs10_pixel}-\ref{tab:context3_vs10_fid_fvd}.

% We examine sensitivity to context quantity and simulation--real outcome mismatch (Figure~\ref{fig:dest_ablation}). For models trained with ten rollouts, one test-time rollout raises
% measured-catalog PSNR by $9.8$--$13.4$\,dB over zero context and stays within $0.8$\,dB of ten.
% VLM-authored results similarly plateau. This limited inference-time gain does not establish that fewer
% training rollouts suffice. With the same three test rollouts, the ten-rollout training configuration exceeds the three-rollout configuration on all six measured-catalog tasks, averaging $0.86$\,dB higher PSNR
% (Table~\ref{tab:context3_vs10_psnr}). Conditioning-channel counts also differ, preventing attribution
% solely to rollout diversity. Other metrics improve on average, with task-specific exceptions (Appendix Tables~\ref{tab:context3_vs10_pixel} and~\ref{tab:context3_vs10_fid_fvd}).

\vspace{-8pt}
\paragraph{Robustness to sim-real outcome mismatch.}

We further stress-test resilience against dynamic discrepancies by deliberately supplying failed simulation rollouts for physically successful demonstrations under identical actions.
In Figure~\ref{fig:dest_ablation}(c,d), At $k=1$, failed simulation context still provides beneficial geometric priors on rigid-body and articulated tasks, yielding measured-catalog $\mathrm{PSNR}$ gains over the unconditioned baseline of $3.1\,\mathrm{dB}$ (\textit{Grasp Cup}), $4.1\,\mathrm{dB}$ (\textit{Open and Close Drawer}), and $3.0\,\mathrm{dB}$ (\textit{Place Cube into Drawer}).
Across VLM-authored tasks, drawer articulation consistently benefits, while deformable tasks (\textit{Wipe Table}) degrade and \textit{Pickup Cup and Cloth} improves as $k$ scales.
More results are in included Appendix Tables~\ref{tab:dest_count_avg}-\ref{tab:fail_k_277}.
These findings suggest that completely failed simulation can still provide useful spatial and physical cues rather than misleading the predictor.
Crucially, our automated Real-to-Sim reconstruction and physical domain randomization pipeline suppresses such simulation false negatives in practice (where physical executions succeed while simulated rollouts fail), ensuring reliable execution guidance.
% \textit{Pickup Cup and Cloth} also benefits, whereas \textit{Wipe Table} and \textit{Remove Cloth} degrade.
% In the VLM-authored setting, \textit{Open and Close Drawer} improves, \textit{Wipe Table} degrades,
% \textit{Grasp Cup} remains near baseline, and \textit{Pickup Cup and Cloth} changes from negative
% to positive gains as $k$ increases. Failed context can therefore remain useful, depending on the task;
% shrinking test subsets and small sample counts limit comparisons across $k$. 

\begin{table}[t]
\centering
\caption{\textbf{Full-rollout PSNR (dB) with matched test context.}
Both models use seeds $0$--$2$, repeated to fill ten
conditioning slots for the ten-rollout model.
Rows denote training $\rightarrow$ test rollout counts.}
\label{tab:context3_vs10_psnr}
\footnotesize
\setlength{\tabcolsep}{3pt}
\resizebox{\linewidth}{!}{
\begin{tabular}{@{}lccccccc@{}}
\toprule
Train $\rightarrow$ Test
% & \shortstack{Wipe\\Table}
% & \shortstack{Grasp\\Cup}
% & \shortstack{Pickup Cup\\and Cloth}
% & \shortstack{Remove\\Cloth}
% & \shortstack{Open and\\Close Drawer}
% & \shortstack{Place Cube\\into Drawer}
& WipeTable
& GraspCup
& PickCup/Cloth
& RemoveCloth
& Open/CloseDrawer
& PlaceCube
& Mean \\
\midrule
$3\rightarrow3$
& 27.04 & 28.03 & 31.89 & 27.12 & 25.72 & 27.14 & 27.82 \\
$10\rightarrow3$
& \textbf{29.04} & \textbf{28.33} & \textbf{32.08}
& \textbf{28.65} & \textbf{26.58} & \textbf{27.43}
& \textbf{28.69} \\
\bottomrule
\end{tabular}
}
% \vspace{-5pt}
\end{table}

\section{Conclusion and Limitations}
\label{sec:limitations_conclusion}
We presented VPTwin, a Real-Sim-Real framework that grounds visual prediction in automated digital twins to suppress physical hallucinations. By conditioning on randomized simulation rollouts and co-training on synthetic counterfactual failures, our approach reliably anticipates candidate action outcomes. As a predictive visual verifier in staged planning, it eliminates false acceptances and improves manipulation success on physical hardware over simulation-only planners.

Several limitations outline avenues for future work. First, automated twin authoring assumes calibrated camera extrinsics and task-specific reconstruction, motivating extensions toward open-world digital twin synthesis. Second, synthetic failure co-training does not yet capture the full diversity of unmodeled real-world dynamics. Finally, while physical trials validate feasibility, scaling to larger multi-stage manipulation suites and closed-loop online adaptation remains promising future work.

% Add your completed AI use and other statements here.
\newpage

\bibliographystyle{iclr2027_conference}
\bibliography{iclr2027_conference}

@inproceedings{he2026neoworld,
  title={NeoWorld-Pro: Programming Interactive Scenes from Monocular Images for Embodied Simulation},
  author={He, Yumeng and Song, Yichen and Yang, Xiaotian and Zhang, Weijia and Zhou, Zanwei and Gong, Junru and Yang, Xiaokang and Wang, Yunbo},
  booktitle={NeurIPS},
  year={2026}
}

@inproceedings{ko2024learning,
  title={Learning to act from actionless videos through dense correspondences},
  author={Ko, Po-Chen and Mao, Jiayuan and Du, Yilun and Sun, Shao-Hua and Tenenbaum, Joshua B},
  booktitle={International Conference on Learning Representations},
  volume={2024},
  pages={40938--40958},
  year={2024}
}

@article{du2023learning,
  title={Learning universal policies via text-guided video generation},
  author={Du, Yilun and Yang, Sherry and Dai, Bo and Dai, Hanjun and Nachum, Ofir and Tenenbaum, Josh and Schuurmans, Dale and Abbeel, Pieter},
  journal={Advances in neural information processing systems},
  volume={36},
  pages={9156--9172},
  year={2023}
}

@article{wu2024ivideogpt,
  title={ivideogpt: Interactive videogpts are scalable world models},
  author={Wu, Jialong and Yin, Shaofeng and Feng, Ningya and He, Xu and Li, Dong and Hao, Jianye and Long, Mingsheng},
  journal={Advances in Neural Information Processing Systems},
  volume={37},
  pages={68082--68119},
  year={2024}
}

@article{zhu2406irasim,
  title={IRASim: A Fine-Grained World Model for Robot Manipulation},
  author={Zhu, Fangqi and Wu, Hongtao and Guo, Song and Liu, Yuxiao and Cheang, Chilam and Kong, Tao},
  journal={https://arxiv. org/abs/2406.14540},
  year={2025}
}

@misc{li2025unifiedvideoactionmodel,
      title={Unified Video Action Model}, 
      author={Shuang Li and Yihuai Gao and Dorsa Sadigh and Shuran Song},
      year={2025},
      eprint={2503.00200},
      archivePrefix={arXiv},
      primaryClass={cs.RO},
      url={https://arxiv.org/abs/2503.00200}, 
}

@article{zhou2024dino,
  title={DINO-WM: World models on pre-trained visual features enable zero-shot planning},
  author={Zhou, Gaoyue and Pan, Hengkai and LeCun, Yann and Pinto, Lerrel},
  journal={arXiv preprint arXiv:2411.04983},
  year={2024}
}

@inproceedings{guo2026ctrl,
  title={Ctrl-World: A controllable generative world model for robot manipulation},
  author={Guo, Yanjiang and Shi, Lucy and Chen, Jianyu and Finn, Chelsea},
  booktitle={International Conference on Learning Representations},
  volume={2026},
  pages={6121--6138},
  year={2026}
}

@article{team2025evaluating,
  title={Evaluating gemini robotics policies in a veo world simulator},
  author={{Gemini Robotics Team} and Choromanski, Krzysztof and Devin, Coline and Du, Yilun and Dwibedi, Debidatta and Gao, Ruiqi and Jindal, Abhishek and Kipf, Thomas and Kirmani, Sean and Leal, Isabel and others},
  journal={arXiv preprint arXiv:2512.10675},
  year={2025}
}

@article{gao2026dreamdojo,
  title={Dreamdojo: A generalist robot world model from large-scale human videos},
  author={Gao, Shenyuan and Liang, William and Zheng, Kaiyuan and Malik, Ayaan and Ye, Seonghyeon and Yu, Sihyun and Tseng, Wei-Cheng and Dong, Yuzhu and Mo, Kaichun and Lin, Chen-Hsuan and others},
  journal={arXiv preprint arXiv:2602.06949},
  year={2026}
}

@article{wang2026interactive,
  title={Interactive world simulator for robot policy training and evaluation},
  author={Wang, Yixuan and Syed, Rhythm and Wu, Fangyu and Zhang, Mengchao and Onol, Aykut and Barreiros, Jose and Nayyeri, Hooshang and Dear, Tony and Zhang, Huan and Li, Yunzhu},
  journal={arXiv preprint arXiv:2603.08546},
  year={2026}
}

@inproceedings{todorov2012mujoco,
  title={Mujoco: A physics engine for model-based control},
  author={Todorov, Emanuel and Erez, Tom and Tassa, Yuval},
  booktitle={2012 IEEE/RSJ international conference on intelligent robots and systems},
  pages={5026--5033},
  year={2012},
  organization={IEEE}
}

@misc{nvidiaisaacsim,
  author       = {{NVIDIA}},
  title        = {{Isaac Sim}},
  howpublished = {Computer software},
  url          = {https://github.com/isaac-sim/IsaacSim},
  note         = {Version 5.1.0-rc.19; accessed September 25, 2026}
}

@article{zhu2020robosuite,
  title={robosuite: A modular simulation framework and benchmark for robot learning},
  author={Zhu, Yuke and Wong, Josiah and Mandlekar, Ajay and Mart{\'\i}n-Mart{\'\i}n, Roberto and Joshi, Abhishek and Lin, Kevin and Maddukuri, Abhiram and Nasiriany, Soroush and Zhu, Yifeng},
  journal={arXiv preprint arXiv:2009.12293},
  year={2020}
}

@inproceedings{fang2025rebot,
  title={Rebot: Scaling robot learning with real-to-sim-to-real robotic video synthesis},
  author={Fang, Yu and Yang, Yue and Zhu, Xinghao and Zheng, Kaiyuan and Bertasius, Gedas and Szafir, Daniel and Ding, Mingyu},
  booktitle={2025 IEEE/RSJ International Conference on Intelligent Robots and Systems (IROS)},
  pages={11351--11358},
  year={2025},
  organization={IEEE}
}

@inproceedings{liu2026robotransfer,
  title={Robotransfer: Controllable geometry-consistent video diffusion for manipulation policy transfer},
  author={Liu, Liu and Wang, Xiaofeng and Zhao, Guosheng and Li, Keyu and Qin, Wenkang and Zhu, Jiagang and Qiu, Jiaxiong and Zhu, Zheng and Huang, Guan and Su, Zhizhong},
  booktitle={Proceedings of the IEEE/CVF Conference on Computer Vision and Pattern Recognition},
  pages={1410--1420},
  year={2026}
}

@inproceedings{chen2026craft,
  title={{CRAFT: Video Diffusion for Bimanual Robot Data Generation}},
  author={Jason Chen and I-Chun Arthur Liu and Gaurav Sukhatme and Daniel Seita},
  booktitle={International Conference on Intelligent Robots and Systems (IROS)},
  Year={2026}
}

@article{ye2025anchordream,
  title={Anchordream: Repurposing video diffusion for embodiment-aware robot data synthesis},
  author={Ye, Junjie and Xue, Rong and Van Hoorick, Basile and Tokmakov, Pavel and Irshad, Muhammad Zubair and Wang, Yue and Guizilini, Vitor},
  journal={arXiv preprint arXiv:2512.11797},
  year={2025}
}

@article{cao2026tract,
  title={TrAct: Bridging Robot Control and Visual Prediction with Visual Tracks},
  author={Cao, Zhi and Ji, Howard and Zhang, Kevin and Ge, Kuangzhi and Li, Fei-Fei and Wu, Jiajun and Huang, Huang},
  journal={arXiv preprint arXiv:2608.24101},
  year={2026}
}

@article{zhang2026robosnap,
  title={RoboSnap: One-Shot Real-to-Sim Scene Generation for Generalizable Robot Learning and Evaluation},
  author={Zhang, Shujie and Yi, Jingkun and Zhong, Weipeng and Zhou, Zirui and Zhu, Yangkun and Wang, Hanqing and Xu, Xudong and Zhang, Weinan and Shen, Chunhua},
  journal={arXiv preprint arXiv:2607.06699},
  year={2026}
}

@article{huang2026soma,
  title={SoMA: A Real-to-Sim Neural Simulator for Robotic Soft-body Manipulation},
  author={Huang, Mu and Wang, Hui and Ren, Kerui and Xu, Linning and Zhou, Yunsong and Yu, Mulin and Dai, Bo and Pang, Jiangmiao},
  journal={arXiv preprint arXiv:2602.02402},
  year={2026}
}

@article{patel2026learning,
  title={Learning Physics-Guided Residual Dynamics for Deformable Object Simulation},
  author={Patel, Shivansh and Zhang, Kaifeng and Pokkali, Sanjay and Lazebnik, Svetlana and Li, Yunzhu},
  journal={arXiv preprint arXiv:2607.13451},
  year={2026}
}

@inproceedings{xiang2025structured,
  title={Structured 3d latents for scalable and versatile 3d generation},
  author={Xiang, Jianfeng and Lv, Zelong and Xu, Sicheng and Deng, Yu and Wang, Ruicheng and Zhang, Bowen and Chen, Dong and Tong, Xin and Yang, Jiaolong},
  booktitle={2025 IEEE/CVF Conference on Computer Vision and Pattern Recognition (CVPR)},
  pages={21469--21480},
  year={2025},
  organization={IEEE}
}

@inproceedings{yang2024holodeck,
  title={Holodeck: Language guided generation of 3d embodied ai environments},
  author={Yang, Yue and Sun, Fan-Yun and Weihs, Luca and VanderBilt, Eli and Herrasti, Alvaro and Han, Winson and Wu, Jiajun and Haber, Nick and Krishna, Ranjay and Liu, Lingjie and others},
  booktitle={2024 IEEE/CVF Conference on Computer Vision and Pattern Recognition (CVPR)},
  pages={16277--16287},
  year={2024},
  organization={IEEE}
}

@inproceedings{pfaff2026scenesmith,
  title={Scenesmith: Agentic generation of simulation-ready indoor scenes},
  author={Pfaff, Nicholas and Cohn, Thomas and Zakharov, Sergey and Cory, Rick and Tedrake, Russ},
  booktitle={Forty-third International Conference on Machine Learning},
  year={2026}
}

@article{zhou2026articraft,
  title={Articraft: An agentic system for scalable articulated 3d asset generation},
  author={Zhou, Matt and Li, Ruining and Lyu, Xiaoyang and Song, Zhaomou and Huang, Zhening and Zheng, Chuanxia and Rupprecht, Christian and Vedaldi, Andrea and Wu, Shangzhe},
  journal={arXiv preprint arXiv:2605.15187},
  year={2026}
}

@article{xiao2026beyond,
  title={Beyond Placement and Articulation: Usage-Driven Code Scenes for Embodied Interaction},
  author={Xiao, Zijian and Ye, Zipeng and Hao, Jinkun and Yang, Xiong and Xie, Yuchen and Yi, Ran},
  journal={arXiv preprint arXiv:2608.18840},
  year={2026}
}

@article{ning2025prompting,
  title={Prompting with the future: Open-world model predictive control with interactive digital twins},
  author={Ning, Chuanruo and Fang, Kuan and Ma, Wei-Chiu},
  journal={arXiv preprint arXiv:2506.13761},
  year={2025}
}

@inproceedings{liu2026simpact,
  title={Simpact: Simulation-enabled action planning using vision-language models},
  author={Liu, Haowen and Yao, Shaoxiong and Chen, Haonan and Gao, Jiawei and Mao, Jiayuan and Huang, Jia-Bin and Du, Yilun},
  booktitle={Proceedings of the IEEE/CVF Conference on Computer Vision and Pattern Recognition},
  pages={20790--20801},
  year={2026}
}

@article{li2026hydra,
  title={Hydra-0: Action Flow for Generalist World Modeling and Control},
  author={Li, Hongyu and Wen, Bowen and Zhu, Xinghao and Wang, Yixuan and Du, Yilun and Li, Yunzhu and Konidaris, George and Birchfield, Stan and Pouya, Soha and Li, Chenran and others},
  journal={arXiv preprint arXiv:2608.18077},
  year={2026}
}

@article{liu2026realwonder,
  title={Realwonder: Real-time physical action-conditioned video generation},
  author={Liu, Wei and Chen, Ziyu and Li, Zizhang and Wang, Yue and Yu, Hong-Xing and Wu, Jiajun},
  journal={arXiv preprint arXiv:2603.05449},
  year={2026}
}

@inproceedings{le2025articulate,
  title={Articulate-anything: Automatic modeling of articulated objects via a vision-language foundation model},
  author={Le, Long and Xie, Jason and Liang, William and Wang, Hung-Ju and Yang, Yue and Ma, Yecheng Jason and Vedder, Kyle and Krishna, Arjun and Jayaraman, Dinesh and Eaton, Eric},
  booktitle={International Conference on Learning Representations},
  volume={2025},
  pages={17578--17602},
  year={2025}
}

@techreport{openai2026gpt6astraapi,
  author      = {{OpenAI}},
  title       = {{GPT-6 Astra System Card}},
  institution = {OpenAI},
  type        = {System card},
  year        = {2026},
  url         = {https://deploymentsafety.openai.com/gpt-6-astra}
}

@book{vanrossum2009python,
  author    = {Van Rossum, Guido and Drake, Fred L.},
  title     = {Python 3 Reference Manual},
  year      = {2009},
  publisher = {CreateSpace},
  address   = {Scotts Valley, CA},
  isbn      = {978-1-4414-1269-0}
}

@misc{cursor2026grok46,
  author       = {{Cursor}},
  title        = {{Cursor Grok 4.6}},
  year         = {2026},
  howpublished = {Cursor Documentation},
  url          = {https://cursor.com/help/models-and-usage/grok-4-6},
  note         = {Accessed: 2026-09-14}
}

@techreport{spacexai2026grok46,
  author      = {{SpaceXAI}},
  title       = {{Model Card: Grok 4.6}},
  institution = {SpaceXAI},
  type        = {Model card},
  year        = {2026},
  url         = {https://media.x.ai/v1/website/card-4p6-4cd2dc57.pdf}
}

@article{chen2026learning,
  title={Learning world models for interactive video generation},
  author={Chen, Taiye and Hu, Xun and Ding, Zihan and Jin, Chi},
  journal={Advances in Neural Information Processing Systems},
  volume={38},
  pages={154456--154483},
  year={2026}
}

@article{chen2024urdformer,
  title={Urdformer: A pipeline for constructing articulated simulation environments from real-world images},
  author={Chen, Zoey and Walsman, Aaron and Memmel, Marius and Mo, Kaichun and Fang, Alex and Vemuri, Karthikeya and Wu, Alan and Fox, Dieter and Gupta, Abhishek},
  journal={arXiv preprint arXiv:2405.11656},
  year={2024}
}

@article{intelligence2025pi_,
  title   = {{$\pi_{0.5}$}: A Vision-Language-Action Model with
             Open-World Generalization},
  author  = {{Physical Intelligence} and Black, Kevin and Brown, Noah
             and Darpinian, James and Dhabalia, Karan and Driess, Danny
             and Esmail, Adnan and Equi, Michael and Finn, Chelsea
             and Fusai, Niccolo and others},
  journal = {arXiv preprint arXiv:2504.16054},
  year    = {2025}
}

@inproceedings{zhang2018unreasonable,
  title={The unreasonable effectiveness of deep features as a perceptual metric},
  author={Zhang, Richard and Isola, Phillip and Efros, Alexei A and Shechtman, Eli and Wang, Oliver},
  booktitle={2018 IEEE/CVF conference on computer vision and pattern recognition},
  pages={586--595},
  year={2018},
  organization={IEEE}
}

@article{heusel2017gans,
  title={{GANs} trained by a two time-scale update rule converge to a local {Nash} equilibrium},
  author={Heusel, Martin and Ramsauer, Hubert and Unterthiner, Thomas and Nessler, Bernhard and Hochreiter, Sepp},
  journal={Advances in neural information processing systems},
  volume={30},
  year={2017}
}

@article{unterthiner2018towards,
  title={Towards accurate generative models of video: A new metric \& challenges},
  author={Unterthiner, Thomas and Van Steenkiste, Sjoerd and Kurach, Karol and Marinier, Raphael and Michalski, Marcin and Gelly, Sylvain},
  journal={arXiv preprint arXiv:1812.01717},
  year={2018}
}

@article{wang2004image,
  title={Image quality assessment: from error visibility to structural similarity},
  author={Wang, Zhou and Bovik, Alan C and Sheikh, Hamid R and Simoncelli, Eero P},
  journal={IEEE transactions on image processing},
  volume={13},
  number={4},
  pages={600--612},
  year={2004},
  publisher={IEEE}
}

@article{wang2002universal,
  title={A universal image quality index},
  author={Wang, Zhou and Bovik, Alan C},
  journal={IEEE signal processing letters},
  volume={9},
  number={3},
  pages={81--84},
  year={2002},
  publisher={IEEE}
}

@article{wang2009mean,
  title={Mean squared error: Love it or leave it? A new look at signal fidelity measures},
  author={Wang, Zhou and Bovik, Alan C},
  journal={IEEE signal processing magazine},
  volume={26},
  number={1},
  pages={98--117},
  year={2009},
  publisher={IEEE}
}

@article{agarwal2026cosmos,
  title={Cosmos 3: Omnimodal world models for physical ai},
  author={Agarwal, Niket and Ali, Arslan and Allen, Jon and Antolini, Martin and Aubame, Adeline and Azzolini, Alisson and Bai, Junjie and Bala, Maciej and Balaji, Yogesh and Bapst, Josh and others},
  journal={arXiv preprint arXiv:2606.02800},
  year={2026}
}

@article{lynch2017modern,
  title={Modern robotics},
  author={Lynch, Kevin M and Park, Frank C},
  journal={Mechanics, Planning, and Control},
  year={2017}
}

@inproceedings{perez2018film,
  title={Film: Visual reasoning with a general conditioning layer},
  author={Perez, Ethan and Strub, Florian and De Vries, Harm and Dumoulin, Vincent and Courville, Aaron},
  booktitle={Proceedings of the AAAI conference on artificial intelligence},
  volume={32},
  number={1},
  year={2018}
}
\clearpage
\appendix
% \section{Appendix}
\clearpage
\appendix

\section*{Appendix}

\section{Experimental Details}
\label{app:experiments}

\subsection{Platform and workspace}
\begin{figure}[!htbp]
    \centering
    \includegraphics[width=0.85\linewidth]{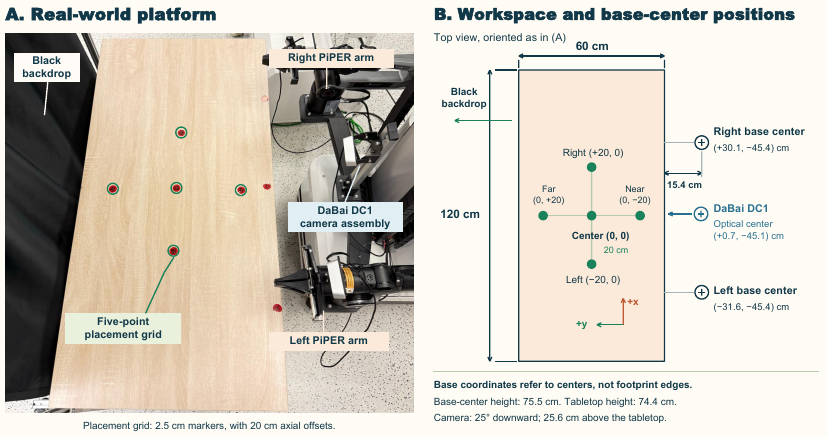}
    \vspace{-8pt}
    \caption{\textbf{Real-robot experimental setup.}
    (A) Bimanual platform, camera, and five-point placement grid.
    (B) Workspace dimensions and base-center coordinates. Each slot is marked by a $2.5$\,cm-diameter red sticker; objects cover the marker without requiring perfect center alignment.}
    \label{fig:experiment-setup}
    \vspace{-8pt}
\end{figure}

The tabletop measures $120\times60$\,cm along the world $x$ and $y$ axes and stands $74.4$\,cm above the floor. The table center defines $(x,y)=(0,0)$, and $+y$ points from the robot toward the black backdrop. Illustrated in Figure~\ref{fig:experiment-setup}. The left and right arm-base centers are at $(-31.6,-45.4)$\,cm and $(+30.1,-45.4)$\,cm, respectively.
Both centers lie $15.4$\,cm beyond the near table edge and $75.5$\,cm above the floor. These coordinates refer to base centers, not footprint edges. The camera optical center is at $(+0.7,-45.1)$\,cm,
$100$\,cm above the floor and $25.6$\,cm above the tabletop. It faces downward at $25^\circ$ from horizontal, with $1.93^\circ$ roll and zero yaw. Its distance to the table-center marker is approximately
$51.9$\,cm.

\subsection{Task placements}
Five red markers, each $2.5$\,cm in diameter, define discrete object placements: center $(0,0)$, left $(-20,0)$, right $(20,0)$, far $(0,20)$, and near $(0,-20)$, in centimeters. Objects occupy these designated points rather than continuously sampled positions. Task instructions and success criteria appear in
Table~\ref{tab:task_def}; placements and object dimensions are listed in Table~\ref{tab:task_placements}.
The markers specify nominal placements rather than exact object centers. An object occupies a slot when its footprint covers the marker; the resulting placement error is not directly measured. Simulation placement perturbations are detailed in Appendix~\ref{app:physics-randomization}; their ranges are not measured bounds on real-world placement error.

\begin{table*}[!b]
\centering
\caption{Placement and object size for the six task families.
The five markers are the table center and four axial offsets of $20$\,cm.
Catalog object dimensions were human-checked; listed masses are nominal coding-agent settings, not physical measurements.
Only \textit{Grasp Cup} and \textit{Open and Close Drawer}
are executed in the real-robot comparison.
The VLM-authored set is Wipe Table, green-cup Grasp Cup,
Pickup Cup and Cloth, and Open and Close Drawer.}
\label{tab:task_placements}
\small
\begin{tabular}{@{}p{0.16\linewidth}p{0.26\linewidth}p{0.28\linewidth}p{0.24\linewidth}@{}}
\toprule
Task & Placement & Catalog reference & VLM suggested \\
\midrule
Wipe Table
& Folded blue towel on one of the five markers.
& Pack $15\times15\times2$\,cm.
& Pack $16\times16\times1.8$\,cm. \\
Grasp Cup
& Upright cup on one of the five markers.
Catalog episodes use a green or blue cup;
VLM episodes use the green cup.
& Height $10$\,cm; bottom / top radius $2.5$ / $3.5$\,cm; $40$\,g.
& Height $10$\,cm; bottom / top radius $2.5$ / $3.48$\,cm; $45$\,g. \\
Pickup Cup and Cloth
& Cloth covers the cup. Both sit on the left, center, or right marker.
& Sheet $50\times53$\,cm, $1.0$\,mm thick, $4$\,g.
& Sheet $52.8\times50.4$\,cm, $0.8$\,mm thick, $4$\,g. \\
Remove Cloth
& Cloth covers the cup. Both sit on the left, center, or right marker.
& Same sheet as Pickup Cup and Cloth.
& Same VLM sheet and VLM cup. This family is catalog-only. \\
Open and Close Drawer
& Fixed on the center marker.
& $20\times20\times7$\,cm; travel $15$\,cm.
& $20\times20\times7.3$\,cm; cavity depth $6.0$\,cm; travel $15.5$\,cm. \\
Place Cube into Drawer
& Fixed at the center; blue or red cube on the left or right marker.
& Cube $5$\,cm, $50$\,g.
& Cube $5$\,cm, $75$\,g. This family is catalog-only. \\
\bottomrule
\end{tabular}
\end{table*}

\subsection{Real-robot evaluation protocol}
The seven \textit{Grasp Cup} trials use center, center, far, near, near, left, and right placements.
The five \textit{Open and Close Drawer} trials use separate initial photographs and executions of the same centered layout. Substage evaluation, refinement, and acceptance are detailed in Appendix~\ref{app:action_planning}.

\section{Scene-asset and Physics-rollout Details}
\label{app:scene-assets}
\subsection{Measured-catalog twin}
\label{app:measured_twin}

The measured catalog specifies the table mesh, camera calibration, robot geometry (The VLM-authored pipeline requires the first three as well), and placement grid described in Appendix~\ref{app:experiments}, but not episode-specific object placements or measured physical properties. Object geometry is human-checked for the 548 episodes across six task families. A coding agent supplies mass, friction, other physical parameters, and their randomization ranges~\citep{cursor2026grok46,spacexai2026grok46}; these properties are not physically measured.

Isaac rollouts each recorded 14-D action trajectory on these assets. Seed~$0$ is the human-accepted rollout. Seeds~$1$--$9$ preserve the recorded actions while randomizing placement, physical parameters, and selected geometric dimensions (Appendix~\ref{app:physics-randomization}). The table, camera, lights, and robot remain fixed across seeds.
\subsection{Physics randomization}
\label{app:physics-randomization}
Placement perturbations act along one tabletop direction, chosen from $\{+x,-x,+y,-y\}$; the other horizontal coordinate and $z$ remain unchanged. Magnitudes are integer millimeters. Measured-catalog seeds $1$--$3$ sample from $\{1,2\}$\,mm, seeds $4$--$6$ from $\{3,4\}$\,mm, and seeds $7$--$9$ use $5$\,mm. VLM-authored seeds sample uniformly from $\{1,2,3,4,5\}$\,mm. Placement perturbation range(1-5mm) suggested by VLM~\citep{openai2026gpt6astraapi}.

\paragraph{Measured-catalog parameters.}
Seeds $1$--$3$, $4$--$6$, and $7$--$9$ use mild, mid, and harsh parameter ranges, respectively
(Table~\ref{tab:physics_ranges}). Interval-valued parameters are sampled uniformly; braced entries denote two-point distributions.

Towel pack height is perturbed by $\delta_h\in\{-3,-2,-1,1,2,3\}$\,mm and clipped to $[1.7,2.3]$\,cm. Cup wall thickness is perturbed by $\delta_w\in\{-1,0,1\}$\,mm and clipped to $[0.3,1.5]$\,mm. These geometric perturbations use the same rule across the three severity levels. Cloth thickness changes only when the seed modulo $9$ belongs to $\{0,3,6\}$, using signed offsets of $1$--$3$\,mm and clipping to $[0.5,3.8]$\,mm.
Ranges suggested by coding agent~\citep{spacexai2026grok46,cursor2026grok46}

\begin{table}[t]
\centering
\caption{Measured-catalog parameter randomization.
Severity levels correspond to seeds $1$--$3$, $4$--$6$,
and $7$--$9$. Friction coefficients are dimensionless. Ranges suggested by coding agent.}
\label{tab:physics_ranges}
\small
\begin{tabular}{@{}lccc@{}}
\toprule
Parameter & Mild & Mid & Harsh \\
\midrule
Towel mass (g)         & $30$--$50$ & $22$--$65$ & $\{20,80\}$ \\
Towel friction         & $0.8$--$1.2$ & $0.55$--$1.4$ & $\{0.40,1.60\}$ \\
Cup mass (g)           & $25$--$48$ & $15$--$55$ & $\{10,60\}$ \\
Cup friction           & $0.9$--$1.5$ & $0.6$--$1.8$ & $\{0.40,2.00\}$ \\
Cloth mass (g)         & $3$--$6$ & $2.5$--$8$ & $\{2,10\}$ \\
Cloth friction         & $1.4$--$2.0$ & $1.0$--$2.1$ & $\{0.60,2.20\}$ \\
Drawer static effort (N)
                       & $1.2$--$3.0$ & $0.7$--$5.0$ & $\{0.40,8.0\}$ \\
Cube mass (g)          & $40$--$70$ & $30$--$90$ & $\{25,120\}$ \\
Cube friction          & $1.2$--$1.8$ & $0.8$--$2.0$ & $\{0.60,2.20\}$ \\
\bottomrule
\end{tabular}
\end{table}

\begin{table*}[!htbp]
\centering
\caption{VLM-suggested parameter ranges.
Each interval is $[\mathrm{suggested}-\mathrm{minus},\,\mathrm{suggested}+\mathrm{plus}]$
from the VLM range file. The green and blue cups share one range set; the red and blue cubes share another.}
\label{tab:vlm_ranges}
\small
\begin{tabular}{@{}lcc@{}}
\toprule
Parameter & Suggested & Suggested range \\
\midrule
Towel width $x$ (cm)        & $16$    & $14$--$18.5$ \\
Towel depth $y$ (cm)        & $16$    & $14$--$19$ \\
Towel pack height (cm)      & $1.8$   & $1.2$--$2.6$ \\
Towel density (kg/m$^3$)    & $90$    & $10$--$210$ \\
Towel mass (g)              & $41.47$ & $1.47$--$101.47$ \\
Towel friction              & $1.05$  & $0.65$--$1.25$ \\
\midrule
Cloth width $x$ (cm)        & $52.8$  & $46.8$--$62.8$ \\
Cloth depth $y$ (cm)        & $50.4$  & $44.4$--$58.4$ \\
Cloth thickness (mm)        & $0.80$  & $0.45$--$1.05$ \\
Cloth areal density (kg/m$^2$) & $0.01509$ & $0.01509$--$0.12$ \\
Cloth friction              & $1.85$  & $0.35$--$2.05$ \\
\midrule
Cup height (cm)             & $10$    & $8.8$--$11.5$ \\
Cup bottom radius (cm)      & $2.5$   & $2.2$--$2.9$ \\
Cup top radius (cm)         & $3.48$  & $3.18$--$3.88$ \\
Cup wall thickness (mm)     & $0.50$  & $0.30$--$1.30$ \\
Cup base thickness (mm)     & $1.0$   & $0.50$--$2.50$ \\
Cup mass (g)                & $45$    & $30$--$70$ \\
Cup friction                & $1.2$   & $0.40$--$1.40$ \\
\midrule
Drawer width $x$ (cm)       & $20$    & $18$--$22.5$ \\
Drawer depth $y$ (cm)       & $20$    & $17.5$--$23$ \\
Drawer height (cm)          & $7.3$   & $6.1$--$8.8$ \\
Drawer cavity depth (cm)    & $6.0$   & $5.0$--$7.2$ \\
Drawer lid well (mm)        & $8$     & $6$--$11$ \\
Drawer outer wall (mm)      & $3$     & $2$--$4$ \\
Drawer tray wall (mm)       & $4$     & $3$--$5$ \\
Drawer side gap (mm)        & $1.0$   & $0.7$--$1.5$ \\
Drawer bottom gap (mm)      & $1.0$   & $0.7$--$1.5$ \\
Drawer front thickness (mm) & $8$     & $6$--$10$ \\
Drawer handle radius (cm)   & $1.4$   & $1.1$--$1.7$ \\
Drawer travel (cm)          & $15.5$  & $13$--$18$ \\
Drawer static effort (N)    & $2.0$   & $1.0$--$3.5$ \\
Drawer dynamic effort (N)   & $1.0$   & $0.55$--$2.0$ \\
Drawer mass (g)             & $400$   & $200$--$700$ \\
Drawer friction             & $0.85$  & $0.45$--$1.05$ \\
\midrule
Cube edge (cm)              & $5.0$   & $4.4$--$5.8$ \\
Cube mass (g)               & $75$    & $30$--$130$ \\
Cube friction               & $1.6$   & $0.45$--$1.80$ \\
\bottomrule
\end{tabular}
\end{table*}

\paragraph{VLM-authored parameters.}
Each parameter is sampled uniformly from its authored interval $[c-\delta_-,c+\delta_+]$, where $c$ is the suggested value and $\delta_-$ and $\delta_+$ are the authored offsets.

\subsection{VLM twin authoring and filtering}
\label{app:authoring}
\begin{figure}[!htbp]
    \centering
    \includegraphics[width=\linewidth]{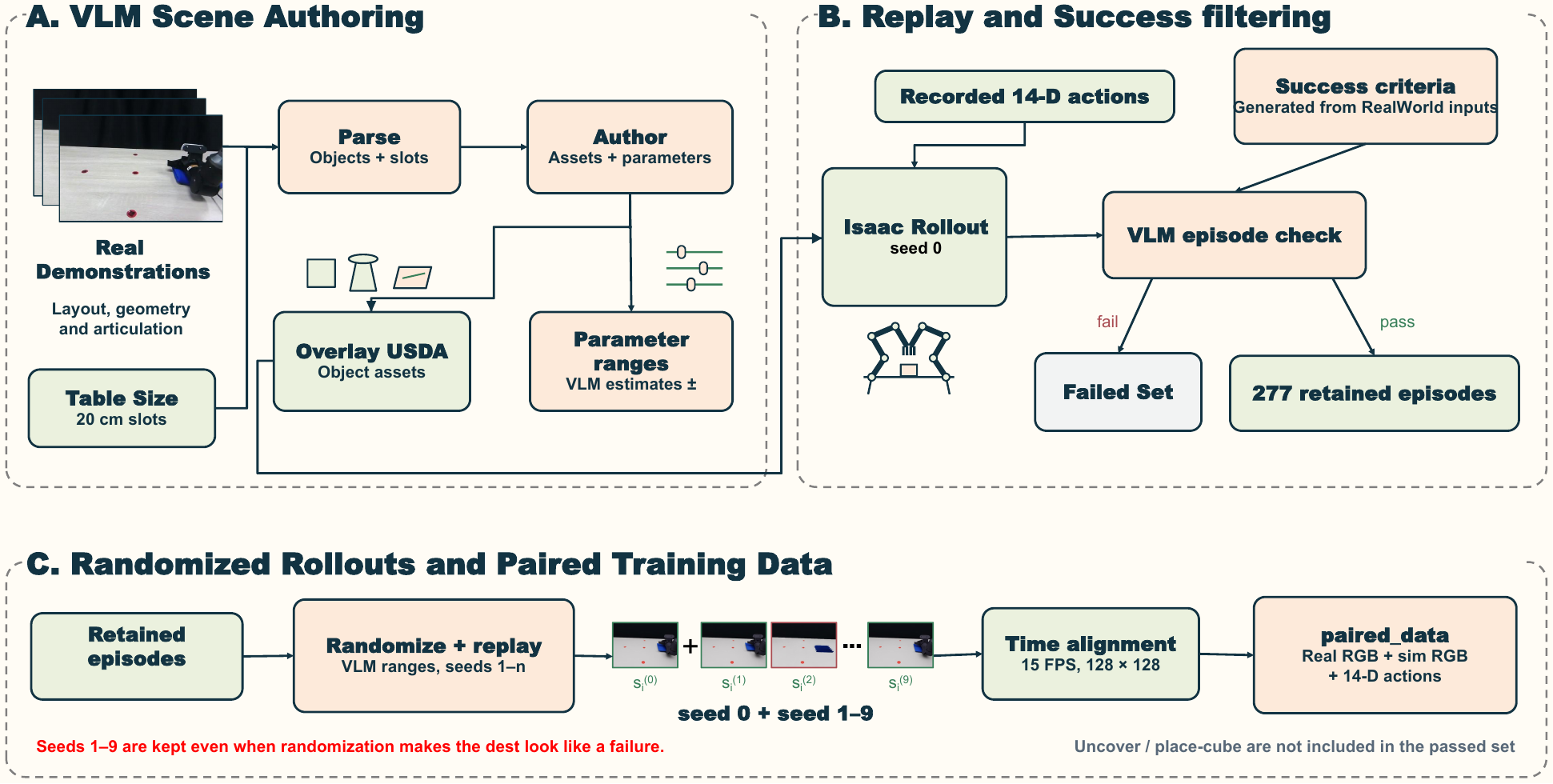}
    \vspace{-8pt}
    \caption{\textbf{VLM scene authoring and simulation-data construction.}
    (A) A VLM authors digital-twin assets and parameter ranges from real
    demonstrations. (B) VLM-derived success criteria retain 277 seed-$0$
    rollouts. (C) Retained episodes are randomized, time-aligned, and
    paired with real RGB and actions for video-predictor training.}
    \label{fig:scene_authoring}
    \vspace{-8pt}
\end{figure}

GPT-6 Astra~\citep{openai2026gpt6astraapi} is called twice: first to parse the scene and then to author the assets. Both calls receive the $0\%$ third-view layout image and later stills at $20$, $40$, $60$, $70$, $80$, and $100\%$. Later stills are used to infer object identity, occlusion, articulation, and motion; the
placement slot is determined from the $0\%$ image. The prompt provides the measured table extents and five-point $20\,\mathrm{cm}$ slot grid. The VLM selects a discrete slot, which the host converts to metric coordinates. The authoring call writes Python~\citep{vanrossum2009python} scripts specifying geometry,
collision shapes, mass, friction, articulation, and sampling ranges. The scripts are executed locally to produce USD assets. The VLM-authored drawer and deformable assets require simulator-specific adaptation. A coding agent converts the drawer into a fixed-base articulation and deploys the cloth and towel as PhysX surface deformables~\citep{cursor2026grok46,spacexai2026grok46, nvidiaisaacsim}. The authoring procedure examines 24 initial-layout variants. The VLM-authored video-prediction set uses 14 of these variants:
five wipe-table, five green-cup grasp, three pickup, and one    drawer variant. The overview pipeline is in Figure~\ref{fig:scene_authoring}.

\paragraph{Success filtering.}
For each task family, the VLM receives real third-view stills at $0$, $20$, $40$, $60$, $80$, and $100\%$ of the episode and writes a textual task instruction and success criterion (Table~\ref{tab:task_def}). Each seed-$0$ dest is then evaluated against that criterion using stills at $0$, $50$, and $100\%$.
Grasp episodes are rechecked on a denser mid-timeline, at $60$, $70$, $75$, and $80\%$ as well as $0$, $50$, and $100\%$. Only accepted seed-$0$ dests enter the VLM video-prediction set, yielding 277 episodes across wipe table, grasp cup, pickup cloth and cup, and open and close drawer.

\section{Fail-dest construction}
\label{app:fail-dests}
The real dataset contains successful demonstrations but no real failure videos. We construct 118 failure episodes from the filtered VLM set: 32 grasp, 43 drawer never-open, and 43 drawer fail-close cases.

\paragraph{Perturbation onset and constraints.}
Grasp and never-open cases preserve the demonstration until five frames before the first gripper-open event.
Fail-close cases preserve the initial pull-and-release sequence and begin perturbation five frames after release. Perturbations affect the active arm's six joints and gripper; some drawer modes hold the other arm fixed from the sign-flip frame onward. The resulting 14-D joint trajectory is clipped to URDF limits. Fingertip targets are constrained to $x\in[-0.22,0.22]$\,m, $z\in[0.78,1.28]$\,m, and the original
trajectory's $y$ range expanded by $0.08$\,m at each end.

\paragraph{Action perturbations.}
Remaining Cartesian displacements receive axis signs $s_x,s_y,s_z\in\{-1,+1\}$, with $\Pr(s_x=-1)=\Pr(s_z=-1)=0.7$ and $\Pr(s_y=+1)=0.7$. If $s_x=s_z=+1$, we set $s_x=-1$. Displacements are scaled by a factor sampled from $U(0.75,1.25)$ and supplemented with a wandering component of amplitude $U(2,6)$\,cm and period $U(12,28)$ frames. Gripper commands use inversion, held-open, held-closed,
or periodic toggling with probabilities $0.25$, $0.20$, $0.20$, and $0.35$, respectively. Toggling intervals range from 6 to 16 frames.

\paragraph{Outcome-based verification.}
Failure labels depend on object trajectories rather than perturbation magnitude. Successful grasp requires a lift of at least $5$\,cm, return to the tabletop, an upright orientation, and stability over the final eight frames. Drawer success requires opening by at least $8$\,cm and subsequently closing to within $3$\,cm. Never-open cases fail the opening condition; fail-close cases satisfy opening but fail closing.

\paragraph{Simulation supervision.}
Isaac Sim rollout s each perturbed action trajectory on the corresponding VLM-authored assets. Seed~$0$ supplies the observation and prediction target; seeds~$1$--$10$ use the same actions with randomized simulation parameters and supply ten conditioning rollouts. Seed~$0$ is excluded from the conditioning set.

%\paragraph{Additional failures outside co-training.}
%Separately generated wipe failures use two recipes. Miss-contact keeps the fingertip at least $20.6$\,cm above
%the tabletop, with joint-step standard deviations $(0.14,0.12,0.12,0.18,0.18,0.16)$\,rad. One-direction %wiping moves $18$--$28$\,cm along one of $\{\pm x,\pm y\}$ after first contact, without a return stroke.
%Miss-contact is verified by the fingertip never entering the region within $8$\,cm above the tabletop.
%These wipe failures and the separately generated pickup failures are excluded from the 118-pair co-training set.

\section{Video Prediction Implementation Details}
\label{app:training}

\subsection{Training settings}
Both stages use AdamW with learning rate $8\times10^{-5}$, weight decay $10^{-4}$, and betas $(0.9,0.999)$.
The learning rate warms up linearly over $10^4$ steps from $10^{-4}$ of its base value. Gradients are clipped at norm $1$, and training uses FP32. Each stage runs for $1{,}000{,}005$ steps; reported predictions use \texttt{final.ckpt}. Stage~1 trains $E$ and $D$ with batch size $1$ and horizon $1$. Stage~2 freezes both and trains $F_\psi$ with batch size $1$ and horizon $H=10$. The action-only baseline uses the same
optimizer, batch size, training budget, frozen autoencoder, data split, and action windows, but omits $C_{i,t}$. Models are trained separately for each task family and arm. The comparisons use five independent training seeds.

\subsection{Stage-1 encoder and decoder}
$E$ consists of Conv2d layers; $D$ combines a consistency U-Net with a ControlNet conditioned on $z=E(r)$.
Stage~1 jointly trains $E$ and $D$ to reconstruct observation frames by denoising RGB conditioned on the encoded latent. Both are frozen in Stage~2. For failure-aware co-training, Stage~1 reconstructs real success
frames and seed-$0$ simulated failure frames; seeds~$1$--$10$ serve only as conditioning rollouts in Stage~2.

\subsection{Aligned Stage-2 windows}

Stage~2 uses $H=10$ time-aligned frames sampled at $15$ Hz and resized
to $128\times128$. We define,
\begin{equation}
s_i^{(k)}[t:t+H]
=
\left(
s_{i,t}^{(k)},\ldots,s_{i,t+H-1}^{(k)}
\right),
\label{eq:appendix-window}
\end{equation}
so the window contains exactly $H$ frames and excludes $t+H$. The same temporal indices are applied to the observation video, 14-D actions, and every dest rollout. Because $E$ operates framewise, $B_{i,t}^{(k)}$ is an $H$-frame latent clip. Success observations are real RGB frames, conditioned on simulation seeds $0$--$9$. Failure observations are seed-$0$ renders under the perturbed action trajectory; conditioning
slot $k\in\{0,\ldots,9\}$ uses seed $k+1$. Observation, action, and conditioning windows use matching
temporal indices.

\subsection{Terminal-only supervision}
Stage~2 processes an $H$-frame latent clip in which the first $H-1$ positions provide the observation history and the final position is the prediction target. For each history position $j<H-1$, the input and target noise levels are matched, $\tau_j=\rho_j$, so the consistency model applies an identity mapping and the corresponding regression residual is zero. Only the terminal position is denoised from noise step
$\tau_{H-1}=999$ to $\rho_{H-1}=0$. Consequently, although the implementation averages the regression loss over all temporal positions, the effective objective is
\begin{equation}
\mathcal{L}_{\mathrm{dyn}}
=
\frac{1}{H}
\left\|
\widehat z_{i,t}^{\ell}
-
z_{i,t}^{\ell}
\right\|_2^2,
\label{eq:appendix-terminal-loss}
\end{equation}
which supervises only the terminal-frame latent. The factor $1/H$ is
constant and does not change the optimization objective.

\subsection{Autoregressive inference}
\label{app:inference}
Inference starts from one encoded observation without padding, using a shorter prefix until the window reaches $H=10$. Each step appends a standard Gaussian latent clipped to $[-6,6]$ and retains at most $H$ slots. Existing slots use stabilization noise index $15$; $F_\psi$ denoises the new slot in one consistency step, $999\rightarrow0$ (\texttt{dyn\_infer\_steps}=1). The updated window, including rewritten history latents, is carried into the next step. The same Stage-1 decoder $D$ renders each latent in three
consistency steps, $999\rightarrow666\rightarrow333\rightarrow0$ (\texttt{dec\_infer\_steps}=3). The configured 50-step DDIM schedule is unused.

\subsection{Episode splits}
\label{app:episode-splits}
Splits are assigned to complete episodes before extracting
training windows. For the measured-catalog set, train,
validation, and test counts are fixed within each task variant
(placement, cup color, or cube color), and episodes are shuffled
within each variant using seed $42$. This yields
$440/54/54$ train/validation/test episodes.

The VLM-authored set inherits these assignments by retaining
the same episode keys that pass VLM filtering, without
reshuffling. It contains $224/26/27$ episodes and excludes
\textit{Remove Cloth} and \textit{Place Cube into Drawer}.
Table~\ref{tab:episode_splits} lists the family-level counts.

\begin{table}[t]
\centering
\caption{Episode-level train/validation/test splits.
The VLM-authored set inherits the measured-catalog assignments.}
\label{tab:episode_splits}
\small
\begin{tabular}{@{}lrrrrrr@{}}
\toprule
& \multicolumn{3}{c}{Measured-catalog}
& \multicolumn{3}{c}{VLM-authored} \\
\cmidrule(lr){2-4} \cmidrule(lr){5-7}
Task & Train & Val & Test & Train & Val & Test \\
\midrule
Wipe Table             & 86  & 10 & 10 & 62 & 8 & 8 \\
Grasp Cup              & 103 & 13 & 13 & 24 & 5 & 3 \\
Pickup Cup and Cloth   & 85  & 10 & 10 & 68 & 5 & 8 \\
Remove Cloth           & 35  & 5  & 5  & -- & -- & -- \\
Open and Close Drawer  & 84  & 10 & 10 & 70 & 8 & 8 \\
Place Cube into Drawer & 47  & 6  & 6  & -- & -- & -- \\
\midrule
Total & 440 & 54 & 54 & 224 & 26 & 27 \\
\bottomrule
\end{tabular}
\end{table}

\subsection{Success--failure split and sampling}
\label{app:failmix-split}
The main video-prediction comparison uses success-only training. Failure-aware co-training is evaluated separately and provides the frozen predictor used for action planning. The co-training set comprises 118 real success episodes and their 118 simulated failure counterparts, whereas the success-only comparator uses the 277-episode success set. Each failure episode inherits its successful parent's split. The 118 success--failure pairs yield 188/26/22 train/validation/test episodes: 48/10/6 for \textit{Grasp Cup} and 140/16/16 for \textit{Open and Close Drawer}. The test set contains three grasp pairs and eight drawer pairs. Training uses balanced sampling of real successes and constructed simulation failures. Success episodes use real RGB targets with seeds~$0$--$9$ as simulation conditions, whereas fail episodes use seed~$0$ as the observation and target and seeds~$1$--$10$ as conditions. The paired set contains 32 grasp episodes and 86 drawer episodes: 43 never-open and 43 fail-close cases. Separately generated failures for 78 wipe and 81 pickup episodes are excluded from this co-training set.

\paragraph{Failure-rollout evaluation starts.}
\textit{First frame} initializes prediction from frame $0$ of the failed seed-$0$ simulation. \textit{Perturbed frame} initializes prediction from the frame at which action perturbation begins. For either start frame $t_0$, the ground-truth seed-$0$ video, perturbed 14-D actions, and ten conditioning rollouts (seeds~$1$--$10$) are all indexed from the same $t_0$. The 30-frame score uses frames $t_0,\ldots,t_0+29$; the full-rollout score uses frames from $t_0$ to the end of that failure episode.

\begin{table*}[t]
\centering
\small
\caption{Definition of tasks. For each manipulation task, we list the corresponding instruction and success criteria. Instruction and success conditions are GPT-6 Astra text authored from multi-time real-video stills. The \textit{Grasp Cup} row describes the return-near-start variant retained in the VLM-authored set; the measured-catalog set also includes cup-relocation variants.}
\label{tab:task_def}
\begin{tabular}{@{}p{0.18\linewidth}p{0.30\linewidth}p{0.46\linewidth}@{}}
\toprule
\textbf{Task} & \textbf{Instruction} & \textbf{Success Condition} \\
\midrule
Remove Cloth
& Remove a white cloth from over a cup while leaving the cup standing on the table. The arm grips and gathers the fabric beside the covered form, then pulls or lifts the cloth clear rather than carrying away the covered object.
& This kind is successful when the arm removes the covering cloth completely, exposing the green cup and leaving it standing on its base near the previously covered location. The cloth must no longer cover or hang over the cup at the end, and the cup must remain on the table rather than leave with the cloth. The final view should show the uncovered cup, the surrounding bare tabletop, and the unchanged red markers after the arm withdraws. \\
\midrule
Place Cube into Drawer
& Open a white drawer using its yellow front knob, pick up the cube from the table, place it inside the open compartment, and close the drawer.
& This kind is successful when the drawer is pulled open far enough to receive the cube, the arm transfers the cube from the tabletop into the compartment, and the drawer is then pushed fully closed. The cube should no longer be at its original outside location or sitting on the unit's top. The final view should show the closed drawer unit standing in place, with the cube enclosed following the visible placement sequence and the surrounding table markers undisturbed. \\
\midrule
Open and Close Drawer
& Pull the initially closed drawer outward to expose its empty compartment, then push it closed again without transferring another object.
& This kind is successful when the arm produces a visible opening of the drawer, revealing the empty compartment, and subsequently returns the drawer front to its closed position against the housing. No object needs to be added or removed. At the end, the unit should remain standing in approximately its original place, the drawer should no longer be extended, and the surrounding tabletop should be unchanged. \\
\midrule
Pickup Cup and Cloth
& Grasp the raised portion of a white cloth-covered bundle and lift the entire bundle away, leaving no uncovered object behind. Its concealed contents are not separately revealed in these photographs.
& This kind is successful when the arm takes hold of the cloth-covered raised portion and removes the whole bundle from the tabletop, including the trailing fabric. The former covered area should be clear at the end, with no cup or other loose object left standing there and no cloth remaining on the table. The visible endpoint is a bare tabletop apart from its stationary red markers; identifying the concealed contents is not necessary to judge that outcome. \\
\midrule
Wipe Table
& Press a blue cloth against the table with the gripper held nearly horizontal and slide it through short local strokes, then lift the arm away while leaving the cloth on the table.
& This kind is successful when the arm maintains contact that presses the blue cloth against the tabletop and moves the cloth through a visible short wiping stroke or back-and-forth motion. It should then withdraw, leaving the cloth resting on the table in roughly the same local area rather than carrying it away. Small shifts, changes in outline, and rearranged folded edges are consistent with the photographs, while the fixed markers remain undisturbed. The supported success criterion is the contact-and-sliding action, not a claim that unseen dirt has been removed. \\
\midrule
Grasp Cup
& Grasp a standing cup, lift it off the table, and set it down near its original position.
& This kind is successful when the gripper takes hold of the cup, lifts it clear of the tabletop, and subsequently returns it to a stable standing position near where it began. The cup may tilt while held or temporarily leave the image, but it should finish upright on its base after the gripper releases and withdraws. The final table arrangement should remain essentially the starting arrangement, apart from a small placement difference, with the red markers unchanged. \\
\bottomrule
\end{tabular}
\end{table*}

\section{Details on Real-Sim-Real Model Predictive Control}
\label{app:action_planning}

This appendix specifies the substage planner of Sec.~\ref{sec:method-planning}.
The motion note is cached text per family, reused for every test $I_0$, and is not a trained motion model.
Items marked \emph{(baseline)} are also given when the whole-skill baseline is evaluated.
The cached motion note, the teleop strips, the six substages, and the video predictor are not.

\paragraph{Cached motion note.}
A VLM reads third-view teleoperation stills of the same task from the predictor's training set, at $0\%,5\%,\ldots,100\%$ (21 frames per episode): five grasp episodes/variations, and three drawer episodes.
The test $I_0$ is a different recording.
The first VLM pass writes how the hand moved, jaw timing, contact, withdrawal, and ordered Cartesian waypoint rules, plus one planner reminder.
The reminder is a single sentence: what happens physically if those waypoints are skipped and one straight MOVE is sent into the object.
For example, for the grasp, the wrist is still parked, the fingers meet the rim or the wall, and the cup is pushed or tipped.
JUDGE scores only the per-substage success/fail test written by the second pass.
The prompt fixes the camera frame ($+x$ image right, $+y$ toward the backdrop, $+z$ up, table top $z=0.744\,\mathrm{m}$) \emph{(baseline)}.
In that frame the model writes ordered waypoint rules: clearance, then jaw gap, then wrist relative to the object.
They contain no joint angles \emph{(baseline)} and no substage list.
The second VLM pass cuts the skill into six substages, each with an allowed primitive subset, a duration hint, ordered waypoints (\texttt{role}, geometric \texttt{where}, jaw, primitive; no metres), and a success/fail test for that substage alone.

\paragraph{Per-substage loop.}
\textbf{RECONSTRUCT} identifies the placement slot from $I_0$ on the five-point grid with $20\,\mathrm{cm}$ axial offsets; the baseline uses its locate call for the same purpose. The drawer occupies the center slot.
Both planners receive slot coordinates and the initial fingertip position in meters in the world coordinate frame. The first substage starts from $I_0$, with the fingertip position computed by forward kinematics from the recorded initial joint configuration $\boldsymbol{\theta}_0$. Later substages use the last committed predicted frame and object and end-effector positions from the committed seed-$0$ simulator state. \textbf{PROPOSE} sees the current image, the matching slot's 21-frame teleop strip, and one seed-$0$ Isaac frame, and writes one short symbolic plan that realises this substage's waypoints in order, using only that substage's allowed action types. The metres in that plan come from the $20\,\mathrm{cm}$ slot on $I_0$ and, after the first substage, from the Isaac state of the committed prefix. \textbf{ACTION-IK} maps the action to 14-D with Piper DH inverse kinematics and $2\,\mathrm{cm}$ linear vias \emph{(baseline executor)}. Every written opening, translation, and wrist field is executed. Isaac Sim rollouts the committed prefix followed by the candidate chunk under seeds $0$--$9$. The predictor uses the current image, candidate action chunk, and ten time-aligned current-substage simulation clips to generate a real-domain future. \textbf{JUDGE} pass/fails this substage from the current image, predicted stills, and the seed-$0$ Isaac window, without teleop frames. A physics miss in Isaac (no contact, no lift, or no opening) fails the substage even if the prediction looks successful. Both need to be satisfied. Predicted stills shown to JUDGE are at $0,20,40,60,80,100\%$ for both grasp and drawer. The seed-$0$ Isaac window is at $0,50,100\%$. The baseline instead scores Isaac seed-$0$ stills/simulation results at $0,20,40,60,80,100\%$ against the VLM-authored whole-skill success text \emph{(baseline)}. \textbf{REFINE}, at most three rounds, rewrites one action from the failed predicted and Isaac stills, the matching-slot teleop cropped to this substage, and the failure history; the revised action is converted by ACTION-IK, simulated, predicted, and judged again.

Motion notes $m$ and substage specifications $\{q_j\}_{j=1}^{J}$ are cached per task family from separate training episodes, with $J=6$. Each $q_j$ defines allowed primitives and a local success criterion. \textsc{RolloutPrefix} rollouts the committed prefix $A$ followed by candidate $u$ from the initial scene under each seed, returning current-substage clips $\mathcal{S}=\{S^{(k)}\}_{k=0}^{9}$ and the terminal seed-$0$ state $s'$. Here, $\Vert$ denotes concatenation. The first passing candidate is committed and refinement stops.
Its action chunk is appended to $A$, and its final predicted frame and simulator state initialize the next substage. If no candidate passes after the initial proposal and up to three refinement rounds, planning terminates without hardware execution. Otherwise, the complete trajectory is executed after all six substages pass both checks (Algorithm~\ref{alg:staged-planning}). Algorithm~\ref{alg:staged-planning} describes normal deployment. For critic evaluation, planning continues through all six substages even if one fails: passing substages use their first passing candidate, and failed substages use the candidate the VLM ranks closest to success. The resulting complete trajectory is executed on hardware while retaining its rejection verdict.

\begin{algorithm}[t]
\caption{Staged action planning through \modelname{}}
\label{alg:staged-planning}
\small
\begin{algorithmic}[1]
\Require VLM $G$, simulator $\mathrm{SIM}$, frozen predictor $P$,
initial image $I_0$, initial joint configuration $\boldsymbol{\theta}_0$,
task $\ell_{\mathrm{task}}$, same-task training demonstrations $\mathcal{D}$
\Ensure Accepted 14-D trajectory $A$ and verdict $b$;
empty trajectory on rejection

\State $(m,\{q_j\}_{j=1}^{J}) \gets$
    \Call{CachedMotionAndStages}{$G,\mathcal{D},\ell_{\mathrm{task}}$}
\State $s_0 \gets$
    \Call{Reconstruct}{$G,I_0,\ell_{\mathrm{task}},\boldsymbol{\theta}_0$}
\State $s \gets s_0$; $I \gets I_0$; $A \gets [\,]$

\For{$j=1,\ldots,J$}
    \State $\mathcal{H} \gets [\,]$; $b_j \gets \mathrm{false}$
    \State $d_j \gets$ \Call{MatchDemo}{$\mathcal{D},q_j,s$}
    \State $\alpha \gets$
        \Call{Propose}{$G,I,d_j,\mathrm{Render}(s),m,q_j$}
    \For{$r=0,\ldots,3$}
        \Comment{Initial proposal plus up to three refinements}
        \State $u \gets$ \Call{ActionIK}{$\alpha$}
        \State $(\mathcal{S},s') \gets$
            \Call{RolloutPrefix}{$\mathrm{SIM},s_0,A,u,\{0,\ldots,9\}$}
        \State $\widehat{V} \gets P(I,u,\mathcal{S})$
        \State $(b_{\mathrm{sim}},b_{\mathrm{vp}},f) \gets$
            \Call{Judge}{$G,q_j,I,S^{(0)},\widehat{V}$}
        \If{$b_{\mathrm{sim}}\land b_{\mathrm{vp}}$}
            \State $b_j \gets \mathrm{true}$; \textbf{break}
        \EndIf
        \State $\mathcal{H} \gets
            \mathcal{H}\mathbin{\Vert}[(\alpha,f)]$
        \If{$r<3$}
            \State $c \gets (u,\widehat{V},S^{(0)},s')$
            \State $\alpha \gets$
                \Call{Refine}{$G,I,d_j,m,q_j,c,\mathcal{H}$}
        \EndIf
    \EndFor
    \If{$\neg b_j$}
        \State \Return $([\,],\mathrm{false})$
            \Comment{No hardware execution}
    \EndIf
    \State $A \gets A\mathbin{\Vert}u$
    \State $I \gets \mathrm{LastFrame}(\widehat{V})$; $s \gets s'$
\EndFor
\State \Call{ExecuteOnRobot}{$A$}
\State \Return $(A,\mathrm{true})$
\end{algorithmic}
\end{algorithm}

\paragraph{Relative to SIMPACT.}
The baseline samples ten whole-skill plans from $I_0$, allows up to three whole-skill revisions, and judges seed-$0$ simulation stills without teleoperation guidance or video prediction. Both planners receive the initial image, metric slot coordinates, initial fingertip position, primitive schema, and whole-skill
success criteria.

\paragraph{Critic agreement.} SIMPACT accepts a plan when a candidate passes its simulation check. VPTwin accepts a plan only when all six substages pass both simulation and video-prediction checks. Accepted plans are deployed normally;  rejected plans are separately executed as described above to measure critic agreement. We report hardware success and critic precision, recall, and F1 using outcomes:
\begin{equation}
\mathrm{Precision}=\frac{\mathrm{TP}}{\mathrm{TP}+\mathrm{FP}},
\quad
\mathrm{Recall}=\frac{\mathrm{TP}}{\mathrm{TP}+\mathrm{FN}},
\quad
\mathrm{F1}=\frac{2\mathrm{TP}}
{2\mathrm{TP}+\mathrm{FP}+\mathrm{FN}}.
\end{equation}
TP (True Positive), FP (False Positive), TN (True Negative), FN (False Negative). TP and FP denote accepted plans that succeed and fail on hardware, respectively; TN and FN denote rejected plans that fail and succeed on hardware, respectively. All rejected plans fail on hardware in this evaluation. The policy $\pi_{0.5}$ has no acceptance decision. We therefore compute precision, recall, and F1 using observed outcomes for both accepted and rejected plans. The policy $\pi_{0.5}$ provides no acceptance decision, so its acceptance and critic metrics are not applicable. For VPTwin on \textit{Grasp Cup}, three of seven planning trials are accepted and executed, and all three succeed, giving an acceptance rate of $3/7$ and precision of $1.00$.

\section{FID and FVD Evaluation Protocol}
\label{app:eval-protocol}
We evaluate full-length test episodes at $128\times128$ resolution. Within each task family, predictions from all test episodes are pooled separately for each model and compared against the same ground-truth set. Baseline and VPTwin use identical test episodes, frame indices, and clip boundaries.

\paragraph{FID.}
FID~\citep{heusel2017gans} uses 2048-dimensional Inception-v3 features extracted with TorchMetrics and its torch-fidelity backend. All frames enter the extractor as uint8 images and are internally resized to $299\times299$.

\paragraph{FVD.}
FVD~\citep{unterthiner2018towards} uses pre-softmax features from the Kinetics-400-pretrained I3D detector distributed with StyleGAN-V in TorchScript format. We extract 16-frame clips with stride 8 within each episode, adding a final clip ending at the last frame if the sliding windows do not cover the episode's end.
Pixel values are scaled to $[-1,1]$, and the detector internally resizes frames to $224\times224$. Clips never cross episode boundaries.

\paragraph{Aggregation and sample counts.}
For each task family, we first average scores over five independent training seeds. Table~\ref{tab:full_episode_avg} reports the mean and standard deviation across these family-level averages: six families in the measured-catalog setting and four in the VLM-authored setting. Table~\ref{tab:fid_fvd_protocol} lists the sample counts shared by both models and the ground-truth reference.
Counts vary across families, and overlapping clips are not independent; comparisons use matched samples within each family.

\paragraph{Context-ablation seeds and aggregation.}
For the dest-count ablation, $k\in\{1,3,5,7,9\}$ uses evaluation seeds $0,1,2$, while $k=0$ uses evaluation seed $0$. The $k=10$ result reuses the existing full-test rollout, without additional evaluation-seed averaging. The failure-only ablation produces one evaluation per $k$. Metrics are computed separately for each task family, checkpoint, and evaluation seed: FID pools the family's frames, and FVD pools its 16-frame clips.

\begin{table}[t]
    \centering
    \caption{Per-family sample counts for FID and FVD.
    $n$ denotes test episodes, $N_{\mathrm{frame}}$ the number
    of frames used for FID, and $N_{\mathrm{clip}}$ the number
    of clips used for FVD.
    Counts are identical for Baseline, VPTwin, and ground truth.}
    \label{tab:fid_fvd_protocol}
    \begin{tabular}{lrrr}
        \toprule
        \textbf{Task family}
        & $n$
        & $N_{\mathrm{frame}}$
        & $N_{\mathrm{clip}}$ \\
        \midrule
        \multicolumn{4}{l}{\textit{Measured-catalog setting}} \\
        Wipe Table             & 10 & 1369 & 166 \\
        Grasp Cup              & 13 & 3346 & 412 \\
        Pickup Cup and Cloth   & 10 & 1207 & 146 \\
        Remove Cloth           &  5 &  713 &  87 \\
        Open and Close Drawer  & 10 & 2463 & 302 \\
        Place Cube into Drawer &  6 & 2471 & 305 \\
        \midrule
        \multicolumn{4}{l}{\textit{VLM-authored setting}} \\
        Wipe Table             & 8 & 1116 & 135 \\
        Grasp Cup              & 3 &  706 &  87 \\
        Pickup Cup and Cloth   & 8 &  943 & 114 \\
        Open and Close Drawer  & 8 & 2004 & 246 \\
        \bottomrule
    \end{tabular}
\end{table}

\begin{table}[t]
    \centering
    \caption{%
        Dest-count ablation, family-averaged (mean $\pm$ std over task families).
        The input retains ten conditioning slots.
        For $k>0$ the $k$ sampled dests are cycled into every slot (no black frames);
        $k{=}0$ is all zeros.
        $k\in\{1,5\}$ averages three evaluation seeds, while $k=10$ uses all ten rollouts.
    }
    \label{tab:dest_count_avg}
    \begin{tabular}{l cccc}
        \toprule
        \textbf{Metric}
        & $\mathbf{k{=}0}$
        & $\mathbf{k{=}1}$
        & $\mathbf{k{=}5}$
        & $\mathbf{k{=}10}$ \\
        \midrule
        \multicolumn{5}{l}{\textit{Measured-catalog (548), 6 families}} \\
        \addlinespace[2pt]
        MSE $\downarrow$
        & $0.101 \pm 0.048$
        & $0.007 \pm 0.002$
        & $0.007 \pm 0.002$
        & $0.007 \pm 0.002$ \\
        LPIPS $\downarrow$
        & $0.579 \pm 0.030$
        & $0.289 \pm 0.055$
        & $0.286 \pm 0.055$
        & $0.284 \pm 0.056$ \\
        FID $\downarrow$
        & $363.3 \pm 18.6$
        & $267.8 \pm 36.5$
        & $266.6 \pm 37.5$
        & $265.6 \pm 36.9$ \\
        PSNR $\uparrow$
        & $16.71 \pm 2.74$
        & $28.36 \pm 1.91$
        & $28.51 \pm 1.89$
        & $28.57 \pm 1.92$ \\
        SSIM $\uparrow$
        & $0.547 \pm 0.110$
        & $0.796 \pm 0.057$
        & $0.797 \pm 0.056$
        & $0.798 \pm 0.058$ \\
        UIQI $\uparrow$
        & $0.053 \pm 0.026$
        & $0.277 \pm 0.040$
        & $0.281 \pm 0.039$
        & $0.283 \pm 0.040$ \\
        FVD $\downarrow$
        & $2884.7 \pm 383.8$
        & $2028.3 \pm 273.2$
        & $2015.7 \pm 262.2$
        & $2014.5 \pm 264.7$ \\
        \midrule
        \multicolumn{5}{l}{\textit{VLM-authored (277), 4 families}} \\
        \addlinespace[2pt]
        MSE $\downarrow$
        & $0.109 \pm 0.059$
        & $0.006 \pm 0.002$
        & $0.007 \pm 0.002$
        & $0.006 \pm 0.002$ \\
        LPIPS $\downarrow$
        & $0.585 \pm 0.051$
        & $0.263 \pm 0.046$
        & $0.264 \pm 0.045$
        & $0.259 \pm 0.048$ \\
        FID $\downarrow$
        & $360.6 \pm 23.3$
        & $261.0 \pm 40.8$
        & $262.1 \pm 38.7$
        & $259.1 \pm 42.3$ \\
        PSNR $\uparrow$
        & $16.78 \pm 4.15$
        & $28.84 \pm 1.77$
        & $28.74 \pm 1.87$
        & $28.96 \pm 1.83$ \\
        SSIM $\uparrow$
        & $0.546 \pm 0.152$
        & $0.807 \pm 0.060$
        & $0.806 \pm 0.060$
        & $0.809 \pm 0.060$ \\
        UIQI $\uparrow$
        & $0.059 \pm 0.042$
        & $0.291 \pm 0.035$
        & $0.291 \pm 0.035$
        & $0.294 \pm 0.039$ \\
        FVD $\downarrow$
        & $3016.2 \pm 724.2$
        & $2091.2 \pm 373.6$
        & $2155.0 \pm 373.7$
        & $2131.0 \pm 340.6$ \\
        \bottomrule
    \end{tabular}
\end{table}
\begin{table*}[t]
    \centering
    \caption{%
        Fail-dest ablation on the measured catalog (548).
        Concat receives the $k$ lowest-index failed dests, cycled into all 10 slots.
        Baseline is the action-only model on the same episodes.
        $k{=}3$ is marked $^{*}$ when $n\le 2$.
        Bold is the better of baseline and concat. $\Delta$ is concat minus baseline.
        $n$ is the $k{=}1$ episode count.
    }
    \label{tab:fail_k_548}
    \footnotesize
    \setlength{\tabcolsep}{4pt}
    \begin{tabular}{ll ccc ccc}
        \toprule
        & & \multicolumn{3}{c}{$\mathbf{k{=}1}$}
        & \multicolumn{3}{c}{$\mathbf{k{=}3}$} \\
        \cmidrule(lr){3-5} \cmidrule(lr){6-8}
        \textbf{Family} & \textbf{Metric}
        & \textbf{B} & \textbf{C} & $\boldsymbol{\Delta}$
        & \textbf{B} & \textbf{C} & $\boldsymbol{\Delta}$ \\
        \midrule
        \multirow{7}{*}{\shortstack[l]{Pickup Cup and Cloth\\$n{=}6$}}
        & MSE $\downarrow$
        & $0.0040$ & $\mathbf{0.0036}$ & $-0.0004$
        & $0.0042$ & $\mathbf{0.0036}$ & $-0.0006$ \\
        & LPIPS $\downarrow$
        & $\mathbf{0.260}$ & $0.277$ & $+0.017$
        & $\mathbf{0.281}$ & $0.293$ & $+0.012$ \\
        & FID $\downarrow$
        & $\mathbf{242.5}$ & $242.8$ & $+0.3$
        & $260.2$ & $\mathbf{254.4}$ & $-5.8$ \\
        & PSNR $\uparrow$
        & $30.28$ & $\mathbf{30.68}$ & $+0.40$
        & $30.05$ & $\mathbf{30.82}$ & $+0.77$ \\
        & SSIM $\uparrow$
        & $\mathbf{0.865}$ & $0.862$ & $-0.003$
        & $0.864$ & $0.864$ & $0.000$ \\
        & UIQI $\uparrow$
        & $\mathbf{0.313}$ & $0.292$ & $-0.021$
        & $\mathbf{0.290}$ & $0.281$ & $-0.009$ \\
        & FVD $\downarrow$
        & $1727.7$ & $\mathbf{1582.3}$ & $-145.4$
        & $1956.3$ & $\mathbf{1875.5}$ & $-80.8$ \\
        \addlinespace[3pt]
        \multirow{7}{*}{\shortstack[l]{Wipe Table\\$n{=}9$}}
        & MSE $\downarrow$
        & $\mathbf{0.0143}$ & $0.0227$ & $+0.0084$
        & $\mathbf{0.0160}$ & $0.0264$ & $+0.0104$ \\
        & LPIPS $\downarrow$
        & $\mathbf{0.257}$ & $0.299$ & $+0.042$
        & $\mathbf{0.252}$ & $0.289$ & $+0.037$ \\
        & FID $\downarrow$
        & $\mathbf{256.4}$ & $301.8$ & $+45.4$
        & $\mathbf{253.9}$ & $294.2$ & $+40.3$ \\
        & PSNR $\uparrow$
        & $\mathbf{25.93}$ & $23.32$ & $-2.62$
        & $\mathbf{25.05}$ & $22.67$ & $-2.37$ \\
        & SSIM $\uparrow$
        & $\mathbf{0.807}$ & $0.800$ & $-0.007$
        & $\mathbf{0.802}$ & $0.794$ & $-0.008$ \\
        & UIQI $\uparrow$
        & $\mathbf{0.293}$ & $0.281$ & $-0.012$
        & $\mathbf{0.291}$ & $0.284$ & $-0.007$ \\
        & FVD $\downarrow$
        & $\mathbf{2316.0}$ & $3066.8$ & $+750.8$
        & $\mathbf{2401.7}$ & $2885.5$ & $+483.8$ \\
        \addlinespace[3pt]
        \multirow{7}{*}{\shortstack[l]{Grasp Cup\\$n{=}5$}}
        & MSE $\downarrow$
        & $0.0140$ & $\mathbf{0.0062}$ & $-0.0078$
        & $0.0168$ & $\mathbf{0.0066}$ & $-0.0102$ \\
        & LPIPS $\downarrow$
        & $0.325$ & $\mathbf{0.293}$ & $-0.032$
        & $0.329$ & $\mathbf{0.288}$ & $-0.041$ \\
        & FID $\downarrow$
        & $297.4$ & $\mathbf{262.7}$ & $-34.7$
        & $307.1$ & $\mathbf{278.2}$ & $-28.9$ \\
        & PSNR $\uparrow$
        & $25.05$ & $\mathbf{28.21}$ & $+3.16$
        & $23.91$ & $\mathbf{27.93}$ & $+4.02$ \\
        & SSIM $\uparrow$
        & $0.765$ & $\mathbf{0.794}$ & $+0.029$
        & $0.757$ & $\mathbf{0.787}$ & $+0.030$ \\
        & UIQI $\uparrow$
        & $0.227$ & $\mathbf{0.267}$ & $+0.040$
        & $0.224$ & $\mathbf{0.264}$ & $+0.040$ \\
        & FVD $\downarrow$
        & $2810.6$ & $\mathbf{2555.0}$ & $-255.6$
        & $2763.8$ & $\mathbf{2615.0}$ & $-148.8$ \\
        \addlinespace[3pt]
        \multirow{7}{*}{\shortstack[l]{Open and Close Drawer\\$n{=}7$}}
        & MSE $\downarrow$
        & $0.0250$ & $\mathbf{0.0092}$ & $-0.0158$
        & $0.0264^{*}$ & $\mathbf{0.0079}^{*}$ & $-0.0185^{*}$ \\
        & LPIPS $\downarrow$
        & $0.391$ & $\mathbf{0.344}$ & $-0.047$
        & $0.386^{*}$ & $\mathbf{0.335}^{*}$ & $-0.051^{*}$ \\
        & FID $\downarrow$
        & $\mathbf{300.7}$ & $306.9$ & $+6.2$
        & $331.5^{*}$ & $\mathbf{326.2}^{*}$ & $-5.3^{*}$ \\
        & PSNR $\uparrow$
        & $22.38$ & $\mathbf{26.46}$ & $+4.08$
        & $21.91^{*}$ & $\mathbf{27.05}^{*}$ & $+5.14^{*}$ \\
        & SSIM $\uparrow$
        & $0.687$ & $\mathbf{0.721}$ & $+0.034$
        & $0.683^{*}$ & $\mathbf{0.727}^{*}$ & $+0.044^{*}$ \\
        & UIQI $\uparrow$
        & $0.204$ & $\mathbf{0.240}$ & $+0.036$
        & $0.200^{*}$ & $\mathbf{0.249}^{*}$ & $+0.049^{*}$ \\
        & FVD $\downarrow$
        & $2111.2$ & $\mathbf{2009.8}$ & $-101.4$
        & $2330.5^{*}$ & $\mathbf{2205.3}^{*}$ & $-125.2^{*}$ \\
        \addlinespace[3pt]
        \multirow{7}{*}{\shortstack[l]{Place Cube into Drawer\\$n{=}4$}}
        & MSE $\downarrow$
        & $0.0181$ & $\mathbf{0.0091}$ & $-0.0090$
        & $0.0179$ & $\mathbf{0.0088}$ & $-0.0091$ \\
        & LPIPS $\downarrow$
        & $0.422$ & $\mathbf{0.360}$ & $-0.062$
        & $0.422$ & $\mathbf{0.357}$ & $-0.065$ \\
        & FID $\downarrow$
        & $\mathbf{299.6}$ & $304.6$ & $+5.0$
        & $\mathbf{303.5}$ & $305.9$ & $+2.4$ \\
        & PSNR $\uparrow$
        & $23.48$ & $\mathbf{26.48}$ & $+3.00$
        & $23.49$ & $\mathbf{26.65}$ & $+3.16$ \\
        & SSIM $\uparrow$
        & $0.710$ & $\mathbf{0.724}$ & $+0.014$
        & $0.707$ & $\mathbf{0.727}$ & $+0.020$ \\
        & UIQI $\uparrow$
        & $0.224$ & $\mathbf{0.236}$ & $+0.012$
        & $0.218$ & $\mathbf{0.239}$ & $+0.021$ \\
        & FVD $\downarrow$
        & $2308.1$ & $\mathbf{2281.2}$ & $-26.9$
        & $\mathbf{2271.4}$ & $2316.7$ & $+45.3$ \\
        \addlinespace[3pt]
        \multirow{7}{*}{\shortstack[l]{Remove Cloth\\$n{=}5$}}
        & MSE $\downarrow$
        & $0.0135$ & $\mathbf{0.0103}$ & $-0.0032$
        & $\mathbf{0.0070}$ & $0.0081$ & $+0.0011$ \\
        & LPIPS $\downarrow$
        & $\mathbf{0.344}$ & $0.359$ & $+0.015$
        & $\mathbf{0.314}$ & $0.331$ & $+0.017$ \\
        & FID $\downarrow$
        & $\mathbf{309.5}$ & $316.0$ & $+6.5$
        & $\mathbf{319.8}$ & $326.1$ & $+6.3$ \\
        & PSNR $\uparrow$
        & $\mathbf{26.43}$ & $26.17$ & $-0.26$
        & $\mathbf{27.69}$ & $27.08$ & $-0.61$ \\
        & SSIM $\uparrow$
        & $\mathbf{0.790}$ & $0.787$ & $-0.003$
        & $\mathbf{0.811}$ & $0.800$ & $-0.011$ \\
        & UIQI $\uparrow$
        & $\mathbf{0.228}$ & $0.204$ & $-0.024$
        & $\mathbf{0.243}$ & $0.218$ & $-0.025$ \\
        & FVD $\downarrow$
        & $\mathbf{1663.2}$ & $1780.6$ & $+117.4$
        & $\mathbf{1526.2}$ & $1567.5$ & $+41.3$ \\
        \bottomrule
    \end{tabular}
\end{table*}

\begin{table*}[t]
    \centering
    \caption{%
        Fail-dest ablation on the VLM-authored set (277).
        Same protocol as Table~\ref{tab:fail_k_548}.
        Grasp Cup is a single episode at $k{=}1$.
        Wipe Table has no episode with three failures.
        Every $k{=}3$ cell that exists is one episode ($^{*}$).
    }
    \label{tab:fail_k_277}
    \footnotesize
    \setlength{\tabcolsep}{4pt}
    \begin{tabular}{ll ccc ccc}
        \toprule
        & & \multicolumn{3}{c}{$\mathbf{k{=}1}$}
        & \multicolumn{3}{c}{$\mathbf{k{=}3}$} \\
        \cmidrule(lr){3-5} \cmidrule(lr){6-8}
        \textbf{Family} & \textbf{Metric}
        & \textbf{B} & \textbf{C} & $\boldsymbol{\Delta}$
        & \textbf{B} & \textbf{C} & $\boldsymbol{\Delta}$ \\
        \midrule
        \multirow{7}{*}{\shortstack[l]{Pickup Cup and Cloth\\$n{=}5$}}
        & MSE $\downarrow$
        & $\mathbf{0.0044}$ & $0.0052$ & $+0.0008$
        & $0.0059^{*}$ & $\mathbf{0.0054}^{*}$ & $-0.0005^{*}$ \\
        & LPIPS $\downarrow$
        & $\mathbf{0.252}$ & $0.339$ & $+0.087$
        & $\mathbf{0.264}^{*}$ & $0.307^{*}$ & $+0.043^{*}$ \\
        & FID $\downarrow$
        & $\mathbf{228.5}$ & $253.1$ & $+24.6$
        & $278.5^{*}$ & $\mathbf{271.5}^{*}$ & $-7.0^{*}$ \\
        & PSNR $\uparrow$
        & $\mathbf{29.82}$ & $29.00$ & $-0.81$
        & $28.29^{*}$ & $\mathbf{28.69}^{*}$ & $+0.41^{*}$ \\
        & SSIM $\uparrow$
        & $\mathbf{0.857}$ & $0.846$ & $-0.011$
        & $0.830^{*}$ & $\mathbf{0.831}^{*}$ & $+0.001^{*}$ \\
        & UIQI $\uparrow$
        & $\mathbf{0.305}$ & $0.248$ & $-0.057$
        & $\mathbf{0.286}^{*}$ & $0.263^{*}$ & $-0.023^{*}$ \\
        & FVD $\downarrow$
        & $1876.4$ & $\mathbf{1478.6}$ & $-397.8$
        & $2383.1^{*}$ & $\mathbf{1838.4}^{*}$ & $-544.7^{*}$ \\
        \addlinespace[3pt]
        \multirow{7}{*}{\shortstack[l]{Wipe Table\\$n{=}6$}}
        & MSE $\downarrow$
        & $\mathbf{0.0211}$ & $0.0279$ & $+0.0068$
        & --- & --- & --- \\
        & LPIPS $\downarrow$
        & $\mathbf{0.279}$ & $0.296$ & $+0.017$
        & --- & --- & --- \\
        & FID $\downarrow$
        & $\mathbf{268.5}$ & $303.4$ & $+34.9$
        & --- & --- & --- \\
        & PSNR $\uparrow$
        & $\mathbf{24.17}$ & $23.04$ & $-1.14$
        & --- & --- & --- \\
        & SSIM $\uparrow$
        & $\mathbf{0.796}$ & $0.792$ & $-0.004$
        & --- & --- & --- \\
        & UIQI $\uparrow$
        & $\mathbf{0.283}$ & $0.278$ & $-0.005$
        & --- & --- & --- \\
        & FVD $\downarrow$
        & $\mathbf{2634.5}$ & $3100.7$ & $+466.2$
        & --- & --- & --- \\
        \addlinespace[3pt]
        \multirow{7}{*}{\shortstack[l]{Grasp Cup\\$n{=}1^{*}$}}
        & MSE $\downarrow$
        & $\mathbf{0.0048}$ & $0.0051$ & $+0.0003$
        & $\mathbf{0.0047}^{*}$ & $0.0050^{*}$ & $+0.0003^{*}$ \\
        & LPIPS $\downarrow$
        & $0.298$ & $\mathbf{0.283}$ & $-0.015$
        & $0.298^{*}$ & $\mathbf{0.282}^{*}$ & $-0.016^{*}$ \\
        & FID $\downarrow$
        & $332.8$ & $\mathbf{310.2}$ & $-22.6$
        & $333.1^{*}$ & $\mathbf{312.3}^{*}$ & $-20.8^{*}$ \\
        & PSNR $\uparrow$
        & $\mathbf{29.25}$ & $28.99$ & $-0.27$
        & $\mathbf{29.27}^{*}$ & $29.03^{*}$ & $-0.24^{*}$ \\
        & SSIM $\uparrow$
        & $0.801$ & $0.801$ & $0.000$
        & $0.801^{*}$ & $0.801^{*}$ & $0.000^{*}$ \\
        & UIQI $\uparrow$
        & $0.248$ & $\mathbf{0.251}$ & $+0.003$
        & $0.247^{*}$ & $\mathbf{0.251}^{*}$ & $+0.004^{*}$ \\
        & FVD $\downarrow$
        & $3462.5$ & $\mathbf{2677.6}$ & $-784.9$
        & $3292.9^{*}$ & $\mathbf{2595.5}^{*}$ & $-697.4^{*}$ \\
        \addlinespace[3pt]
        \multirow{7}{*}{\shortstack[l]{Open and Close Drawer\\$n{=}3$}}
        & MSE $\downarrow$
        & $0.0153$ & $\mathbf{0.0078}$ & $-0.0075$
        & $0.0179^{*}$ & $\mathbf{0.0080}^{*}$ & $-0.0099^{*}$ \\
        & LPIPS $\downarrow$
        & $0.372$ & $\mathbf{0.333}$ & $-0.039$
        & $0.378^{*}$ & $\mathbf{0.348}^{*}$ & $-0.030^{*}$ \\
        & FID $\downarrow$
        & $330.4$ & $\mathbf{320.2}$ & $-10.2$
        & $362.6^{*}$ & $\mathbf{357.4}^{*}$ & $-5.2^{*}$ \\
        & PSNR $\uparrow$
        & $24.20$ & $\mathbf{27.11}$ & $+2.91$
        & $23.50^{*}$ & $\mathbf{26.99}^{*}$ & $+3.48^{*}$ \\
        & SSIM $\uparrow$
        & $0.704$ & $\mathbf{0.728}$ & $+0.024$
        & $0.688^{*}$ & $\mathbf{0.722}^{*}$ & $+0.034^{*}$ \\
        & UIQI $\uparrow$
        & $0.219$ & $\mathbf{0.250}$ & $+0.031$
        & $0.201^{*}$ & $\mathbf{0.237}^{*}$ & $+0.036^{*}$ \\
        & FVD $\downarrow$
        & $\mathbf{2053.0}$ & $2067.2$ & $+14.2$
        & $\mathbf{2388.1}^{*}$ & $2582.2^{*}$ & $+194.1^{*}$ \\
        \bottomrule
    \end{tabular}
\end{table*}

\begin{table*}[t]
    \centering
    \caption{
        Per-family full-episode scores.
        Each cell is Baseline~/-~Ours.
        Baseline is action-only VPTwin; Ours is simulation-conditioned VPTwin (10 dests).
        Pixel metrics are mean $\pm$ std over test episodes.
        Higher PSNR, SSIM, and UIQI are better ($\uparrow$); lower MSE, LPIPS, FID, and FVD
        are better ($\downarrow$). The better value is bolded.
        These family scores are the entries averaged in Table~\ref{tab:full_episode_avg}.
    }
    \label{tab:full_episode_per_family}
    \resizebox{\textwidth}{!}{%
    \begin{tabular}{l c c c c c c c c}
        \toprule
        \textbf{Task family}
        & $\boldsymbol{n}$
        & \textbf{MSE $\downarrow$}
        & \textbf{LPIPS $\downarrow$}
        & \textbf{FID $\downarrow$}
        & \textbf{FVD $\downarrow$}
        & \textbf{PSNR $\uparrow$}
        & \textbf{SSIM $\uparrow$}
        & \textbf{UIQI $\uparrow$} \\
        \midrule
        \multicolumn{9}{l}{\textit{Measured-catalog (548), six task families}} \\
        \addlinespace[2pt]
        Wipe Table & 10
        & $0.0137 \pm 0.0153$ / $\mathbf{0.0063 \pm 0.0050}$
        & $0.252 \pm 0.037$ / $\mathbf{0.213 \pm 0.014}$
        & $250.3$ / $\mathbf{230.6}$
        & $2332.2$ / $\mathbf{2194.8}$
        & $26.25 \pm 3.46$ / $\mathbf{28.93 \pm 2.80}$
        & $0.812 \pm 0.051$ / $\mathbf{0.847 \pm 0.026}$
        & $0.296 \pm 0.033$ / $\mathbf{0.336 \pm 0.016}$ \\
        Grasp Cup & 13
        & $0.0239 \pm 0.0204$ / $\mathbf{0.0067 \pm 0.0032}$
        & $0.349 \pm 0.051$ / $\mathbf{0.277 \pm 0.021}$
        & $278.5$ / $\mathbf{254.2}$
        & $2644.5$ / $\mathbf{2396.9}$
        & $23.54 \pm 3.66$ / $\mathbf{28.26 \pm 2.33}$
        & $0.747 \pm 0.048$ / $\mathbf{0.794 \pm 0.019}$
        & $0.216 \pm 0.029$ / $\mathbf{0.270 \pm 0.013}$ \\
        Pickup Cup and Cloth & 10
        & $0.0036 \pm 0.0012$ / $\mathbf{0.0026 \pm 0.0010}$
        & $0.237 \pm 0.049$ / $\mathbf{0.223 \pm 0.032}$
        & $225.0$ / $\mathbf{216.5}$
        & $\mathbf{1689.7}$ / $1700.6$
        & $30.73 \pm 1.40$ / $\mathbf{32.11 \pm 1.56}$
        & $0.872 \pm 0.019$ / $\mathbf{0.876 \pm 0.021}$
        & $0.324 \pm 0.038$ / $\mathbf{0.329 \pm 0.034}$ \\
        Remove Cloth & 5
        & $0.0132 \pm 0.0162$ / $\mathbf{0.0080 \pm 0.0066}$
        & $0.337 \pm 0.076$ / $\mathbf{0.310 \pm 0.038}$
        & $306.0$ / $\mathbf{295.4}$
        & $\mathbf{1633.7}$ / $1748.7$
        & $26.68 \pm 4.06$ / $\mathbf{28.20 \pm 3.65}$
        & $0.793 \pm 0.061$ / $\mathbf{0.802 \pm 0.049}$
        & $0.234 \pm 0.046$ / $\mathbf{0.240 \pm 0.039}$ \\
        Open and Close Drawer & 10
        & $0.0243 \pm 0.0092$ / $\mathbf{0.0090 \pm 0.0016}$
        & $0.394 \pm 0.028$ / $\mathbf{0.337 \pm 0.010}$
        & $\mathbf{299.0}$ / $300.9$
        & $2147.3$ / $\mathbf{1988.5}$
        & $22.48 \pm 1.78$ / $\mathbf{26.56 \pm 0.75}$
        & $0.687 \pm 0.021$ / $\mathbf{0.728 \pm 0.010}$
        & $0.206 \pm 0.019$ / $\mathbf{0.255 \pm 0.009}$ \\
        Place Cube into Drawer & 6
        & $0.0168 \pm 0.0026$ / $\mathbf{0.0074 \pm 0.0010}$
        & $0.408 \pm 0.025$ / $\mathbf{0.343 \pm 0.009}$
        & $297.2$ / $\mathbf{295.7}$
        & $2122.6$ / $\mathbf{2057.7}$
        & $23.82 \pm 0.68$ / $\mathbf{27.37 \pm 0.59}$
        & $0.707 \pm 0.008$ / $\mathbf{0.741 \pm 0.009}$
        & $0.220 \pm 0.011$ / $\mathbf{0.265 \pm 0.012}$ \\
        \midrule
        \multicolumn{9}{l}{\textit{VLM-authored (277), four task families}} \\
        \addlinespace[2pt]
        Wipe Table & 8
        & $0.0189 \pm 0.0154$ / $\mathbf{0.0069 \pm 0.0064}$
        & $0.279 \pm 0.050$ / $\mathbf{0.215 \pm 0.018}$
        & $265.9$ / $\mathbf{234.0}$
        & $2622.7$ / $\mathbf{2173.8}$
        & $24.47 \pm 3.49$ / $\mathbf{28.91 \pm 3.33}$
        & $0.802 \pm 0.054$ / $\mathbf{0.844 \pm 0.034}$
        & $0.286 \pm 0.035$ / $\mathbf{0.335 \pm 0.024}$ \\
        Grasp Cup & 3
        & $0.0222 \pm 0.0164$ / $\mathbf{0.0072 \pm 0.0035}$
        & $0.347 \pm 0.052$ / $\mathbf{0.268 \pm 0.027}$
        & $311.9$ / $\mathbf{282.6}$
        & $3359.1$ / $\mathbf{2575.7}$
        & $24.22 \pm 5.48$ / $\mathbf{27.91 \pm 2.62}$
        & $0.759 \pm 0.047$ / $\mathbf{0.785 \pm 0.021}$
        & $0.233 \pm 0.028$ / $\mathbf{0.262 \pm 0.016}$ \\
        Pickup Cup and Cloth & 8
        & $0.0037 \pm 0.0015$ / $\mathbf{0.0030 \pm 0.0013}$
        & $0.233 \pm 0.046$ / $\mathbf{0.231 \pm 0.035}$
        & $218.5$ / $\mathbf{214.1}$
        & $1774.6$ / $\mathbf{1765.8}$
        & $30.60 \pm 1.62$ / $\mathbf{31.56 \pm 1.79}$
        & $0.866 \pm 0.022$ / $\mathbf{0.870 \pm 0.023}$
        & $0.317 \pm 0.042$ / $\mathbf{0.320 \pm 0.034}$ \\
        Open and Close Drawer & 8
        & $0.0206 \pm 0.0098$ / $\mathbf{0.0072 \pm 0.0009}$
        & $0.382 \pm 0.028$ / $\mathbf{0.322 \pm 0.014}$
        & $314.4$ / $\mathbf{305.6}$
        & $2043.6$ / $\mathbf{2008.5}$
        & $23.26 \pm 1.88$ / $\mathbf{27.48 \pm 0.56}$
        & $0.692 \pm 0.023$ / $\mathbf{0.736 \pm 0.009}$
        & $0.212 \pm 0.023$ / $\mathbf{0.261 \pm 0.010}$ \\
        \bottomrule
    \end{tabular}%
    }
\end{table*}

\begin{table}[t]
\centering
\caption{Failure-rollout SSIM on the same 11 test pairs as
Table~\ref{tab:failmix_277}. Failmix uses success--failure
co-training; Success-only uses the 277 successful VLM-filtered
episodes. Scores use a 30-frame prefix or the full rollout
from the indicated start frame. Higher is better.}
\label{tab:failmix_277_ssim}
\small
\begin{tabular}{@{}llcccc@{}}
\toprule
& & \multicolumn{2}{c}{30 frames}
  & \multicolumn{2}{c}{Full rollout} \\
\cmidrule(lr){3-4}\cmidrule(lr){5-6}
Start & Task & Failmix & Success-only
             & Failmix & Success-only \\
\midrule
\multirow{2}{*}{First frame}
& Grasp Cup & \textbf{0.714} & 0.704
            & \textbf{0.707} & 0.700 \\
& Open and Close Drawer & \textbf{0.674} & 0.637
                        & \textbf{0.624} & 0.579 \\
\midrule
\multirow{2}{*}{Perturbed frame}
& Grasp Cup & \textbf{0.713} & 0.703
            & \textbf{0.707} & 0.700 \\
& Open and Close Drawer & \textbf{0.620} & 0.590
                        & \textbf{0.605} & 0.558 \\
\bottomrule
\end{tabular}
\end{table}

\begin{table}[t]
\centering
\caption{Full-rollout MSE, SSIM, LPIPS, and UIQI for Table~\ref{tab:context3_vs10_psnr}.
Same episodes and the same two settings.}
\label{tab:context3_vs10_pixel}
\small
\begin{tabular}{lcc cc cc cc}
\toprule
& \multicolumn{2}{c}{\textbf{MSE} $\downarrow$}
& \multicolumn{2}{c}{\textbf{SSIM} $\uparrow$}
& \multicolumn{2}{c}{\textbf{LPIPS} $\downarrow$}
& \multicolumn{2}{c}{\textbf{UIQI} $\uparrow$} \\
\cmidrule(lr){2-3} \cmidrule(lr){4-5} \cmidrule(lr){6-7} \cmidrule(lr){8-9}
\textbf{Task} & 3$\rightarrow$3 & 10$\rightarrow$3
& 3$\rightarrow$3 & 10$\rightarrow$3
& 3$\rightarrow$3 & 10$\rightarrow$3
& 3$\rightarrow$3 & 10$\rightarrow$3 \\
\midrule
Wipe Table & 0.0124 & \textbf{0.0060}
& 0.823 & \textbf{0.847}
& 0.197 & \textbf{0.173}
& 0.312 & \textbf{0.335} \\
Grasp Cup & 0.0073 & \textbf{0.0067}
& 0.792 & \textbf{0.794}
& 0.244 & \textbf{0.241}
& 0.266 & \textbf{0.269} \\
Pickup Cup and Cloth & 0.0028 & \textbf{0.0026}
& 0.876 & 0.876
& \textbf{0.173} & 0.174
& 0.329 & 0.329 \\
Remove Cloth & 0.0117 & \textbf{0.0068}
& 0.800 & \textbf{0.806}
& 0.261 & \textbf{0.252}
& \textbf{0.248} & 0.240 \\
Open and Close Drawer & 0.0112 & \textbf{0.0090}
& 0.723 & \textbf{0.730}
& 0.314 & \textbf{0.304}
& 0.247 & \textbf{0.255} \\
Place Cube into Drawer & 0.0078 & \textbf{0.0073}
& 0.741 & \textbf{0.743}
& 0.315 & \textbf{0.292}
& 0.262 & \textbf{0.268} \\
\midrule
Mean over tasks & 0.0089 & \textbf{0.0064}
& 0.793 & \textbf{0.799}
& 0.251 & \textbf{0.239}
& 0.277 & \textbf{0.283} \\
\bottomrule
\end{tabular}
\end{table}

\begin{table}[t]
\centering
\caption{Pooled FID and FVD for Table~\ref{tab:context3_vs10_psnr}.
FID uses every predicted frame (Inception-2048).
FVD uses 16-frame clips with stride 8.
Both are one score per task, not a mean of per-episode scores.
Lower is better.}
\label{tab:context3_vs10_fid_fvd}
\small
\begin{tabular}{lcc cc}
\toprule
& \multicolumn{2}{c}{\textbf{FID} $\downarrow$}
& \multicolumn{2}{c}{\textbf{FVD} $\downarrow$} \\
\cmidrule(lr){2-3} \cmidrule(lr){4-5}
\textbf{Task} & 3$\rightarrow$3 & 10$\rightarrow$3
& 3$\rightarrow$3 & 10$\rightarrow$3 \\
\midrule
Wipe Table & 231.90 & \textbf{230.12}
& 2182.2 & \textbf{2136.6} \\
Grasp Cup & \textbf{255.10} & 255.43
& 2431.9 & \textbf{2391.0} \\
Pickup Cup and Cloth & 220.18 & \textbf{217.52}
& 1709.4 & \textbf{1693.5} \\
Remove Cloth & 296.15 & \textbf{292.55}
& \textbf{1653.2} & 1781.9 \\
Open and Close Drawer & 309.82 & \textbf{307.23}
& 2087.1 & \textbf{2011.7} \\
Place Cube into Drawer & 294.68 & \textbf{294.55}
& 2091.2 & \textbf{2070.8} \\
\midrule
Mean over tasks & 267.97 & \textbf{266.23}
& 2025.8 & \textbf{2014.2} \\
\bottomrule
\end{tabular}
\end{table}

\section{Prompt Examples}
\label{app:prompts}

The following is the prompt sent to the VLM for scene parsing.

\begin{lstlisting}[
    basicstyle=\ttfamily\footnotesize,
    breaklines=true,
    breakatwhitespace=false,
    columns=fullflexible,
    keepspaces=true,
    showstringspaces=false
]
Look at these photographs. Reason step by step (chain of thought) about what is on the table besides the given table, floor, and black cloth.

For each distinct object: what it looks like; whether anything covers it or sits on it; whether an interior is visible; whether one part can move relative to another. Watch which part stays on the table and which part moves. Do not model the table, floor, black cloth, robot, camera, or the red marks on the tabletop.

Do not assume a catalog. Only write what the photographs support. Do not list the red marks on the tabletop as objects; they are already on the given table. Use the centre mark as the visual origin only.

After each Image 01..24 (0% of that episode) there are later photographs from that same real episode (every 20% through 100%, with additional 70%). They are not layout images. Use them for object identity, covering, interiors, and how parts move. Do not use them to decide where objects sit at 0%. If something is hidden in a 0% photograph and later photographs of that episode reveal it, it still belongs in that 0% layout (under or inside whatever covered it). A later photograph of an empty table is a later moment; it does not mean those objects were absent at 0%. If several 0% photographs show the same kind of covering (the same sheet over a similar raised shape), give the thing underneath one object id even when one episode never uncovers it. Do not invent a second anonymous support for an episode whose later frames go empty. 3 extra photographs come from another episode of the same task; also not layout images.

Write the reasoning in prose first. Then fence one JSON file:

```json scene_parse.json
{
  "objects": [
    {
      "id": "obj_1",
      "appearance": "what it looks like in the photographs",
      "parts": [{"id": "part_1", "appearance": "...", "motion": "stays_put|translates_in_plane|lifts_off|rotates|flexes|unknown"}],
      "softbody": "none|thin_shell|folded_pack",
      "relations": [{"other_id": "obj_2", "relation": "on|under|covering|covered_by|inside|contains|attached_to|moves_relative_to|rests_on"}],
      "notes": "occlusion, interior if visible, how parts move"
    }
  ],
  "per_image": [
    {"image": 1, "visible": ["obj_1"], "present": ["obj_1", "obj_2"]}
  ]
}
````

`image` is 1..24 matching Image 01..24 (the 0% layout photographs). Extra photographs do not get a `per_image` row.
`visible` is what you can see in that 0% photograph.
`present` is what is actually on the table at that 0% layout, including objects only revealed later in the same episode (covered, inside, behind). `present` must include every `visible` id. Do not drop a covered object from `present` because it is missing from `visible`.
If an object flexes, `softbody` is exactly one of two kinds. `thin_shell` = one connected sheet (a drape, cover, or unfolded cloth); its thickness is the fabric thickness. `folded_pack` = a compact folded towel or cloth pack sitting on the table. Do not unfold it into plies and do not describe stacked sheets. A folded pack is approximated later as one soft rectangular block (volume) or one midplane surface whose thickness is the pack height (how tall the pack sits), not a single-ply cloth thickness. Rigid objects use `none`. Do not invent a third kind.
\end{lstlisting}

The following prompt was sent to the VLM to derive task instructions and success criteria from real demonstration stills. Additional information will be released when the repository goes public.

\begin{lstlisting}[
    basicstyle=\ttfamily\footnotesize,
    breaklines=true,
    breakatwhitespace=false,
    columns=fullflexible,
    keepspaces=true,
    showstringspaces=false
]
These photographs are real third-view stills. They are grouped by episode.

Image 01..24 is 0% of that episode. Immediately after each one you will see later stills from that same real episode at every 20% through 100%. Use each group as one timeline.

For each Image 01..24 group, reason step by step about the interaction from 0% to 100%: what starts on the table, what the arm does, what moves or stays, and how the table looks at the end. The arm may appear; describe what it does. Do not reconstruct meshes.

Some groups show the same kind of interaction, only starting at a different place on the table. Merge those into one kind. Do not invent names from a catalog. Refer to groups as Image 01, Image 02, and so on.

Then fence one JSON file. `success` is one paragraph: when this kind of episode would count as successful, written from what the photographs show.

```json success.json
{
  "episodes": [
    {"image": 1, "interaction": "what happens from 0% to 100% in this episode"}
  ],
  "kinds": [
    {
      "id": "kind_1",
      "images": [1, 2],
      "interaction": "what these episodes share",
      "success": "one paragraph: this kind is successful when ..."
    }
  ]
}
```

Every image 1..24 must appear in exactly one kind's `images` list.
\end{lstlisting}

\end{document}